\documentclass[12pt]{article}%
\usepackage{amsmath}
\usepackage{amsfonts}
\usepackage{amssymb}
\usepackage{longtable}
\usepackage{multirow}
\usepackage{rotating}
\usepackage{graphicx}%
\usepackage{color}
\usepackage{graphics}
\usepackage{graphicx}
\usepackage{epsfig}
\usepackage{subfigure}
\usepackage{latexsym,bm}
\usepackage{graphicx}%
\usepackage{mathrsfs}
\usepackage{cite}
\usepackage{mathtools}
\usepackage{url}
\usepackage{hyperref}
\usepackage{diagbox} 
\usepackage{pifont}

\usepackage{caption}  
\usepackage{threeparttable}  

\usepackage{algorithm}
\usepackage{algpseudocode}

\usepackage[usenames,dvipsnames]{xcolor}
\usepackage{tcolorbox}
\usepackage{tabularx}
\usepackage{array}
\usepackage{colortbl}
\tcbuselibrary{skins}

\usepackage[T1]{fontenc}
\usepackage[utf8]{inputenc}

\newtheorem{example}{Example}

\begin{document}

\def\R{{\mathbb R}}
\newcommand{\bbb}[1]{{\boldsymbol  #1 }}
\newcommand{\vth}{{\vartheta}}
\newcommand{\blue}[1]{{\color{blue} #1 }}
\newcommand{\red}[1]{{\color{red} #1 }}
\newcommand{\magenta}[1]{{\color{magenta} #1 }}
\newcommand{\g}[1]{{\color{teal} #1 }}

\newcommand{\srn}{Sqr-ResNet }
\newcommand{\srnd}{Sqr-ResNet. }
\newcommand{\srnc}{Sqr-ResNet, }
\newcommand{\rn}{Hw-net }
\newcommand{\rnc}{Hw-net,}
\newcommand{\rnd}{Hw-net.}

\newcommand{\eff}{$\kappa_{\rm eff}$}
\title{Stable Weight Updating: A Key to Reliable PDE Solutions Using Deep Learning}


\author{
A. Noorizadegan\thanks{Department of Civil Engineering, National Taiwan University, 10617, Taipei, Taiwan}
, R. Cavoretto\thanks{Department of Mathematics \lq\lq Giuseppe Peano\rq\rq, University of Torino, via Carlo Alberto 10, 10123 Torino, Italy}
, D.L. Young \thanks{Core Tech System Co. Ltd, Moldex3D, Chubei, Taiwan} \footnotemark[1]
, C.S. Chen\footnotemark[1] \thanks{Corresponding authors:\quad \texttt{dchen@ntu.edu.tw}} 
}

\date{}

\maketitle

\begin{abstract}
\noindent
\textit{Background:}
Deep learning techniques, particularly neural networks, have revolutionized computational physics, offering powerful tools for solving complex partial differential equations (PDEs). However, ensuring stability and efficiency remains a challenge, especially in scenarios involving nonlinear and time-dependent equations.\\
\textit{Methodology:}
This paper introduces novel residual-based architectures, namely the Simple Highway Network and the Squared Residual Network, designed to enhance stability and accuracy in physics-informed neural networks (PINNs). These architectures augment traditional neural networks by incorporating residual connections, which facilitate smoother weight updates and improve backpropagation efficiency.\\
\textit{Results:}
Through extensive numerical experiments across various examples—includ\-ing linear and nonlinear, time-dependent and independent PDEs—we demonstrate the efficacy of the proposed architectures. The Squared Residual Network, in particular, exhibits robust performance, achieving enhanced stability and accuracy compared to conventional neural networks. These findings underscore the potential of residual-based architectures in advancing deep learning for PDEs and computational physics applications.

\end{abstract}

\section{Introduction}
\label{sec:intro}
Deep neural networks have profoundly impacted machine learning and artificial intelligence, achieving remarkable results in diverse fields such as image recognition, natural language processing, and reinforcement learning. Their combination with physics-informed methodologies has been particularly promising, as highlighted by Raissi et al. \cite{Raissi19}.

However, many works have been done to address the instability in deep learning for solving PDE problems, as highlighted by \cite{Bajaj23, Wang21, Fang23, Wang22a, Wang22}. Approaches to improve the performance of neural networks include optimizing the loss function \cite{Yu22}, updating weights more effectively \cite{McClenny23}, strategically selecting collocation points \cite{Cho24,Hou23}, and modifying the architecture of neural networks \cite{Wang21,Tanyu23}.

Residual networks (ResNets) have revolutionized deep learning, particularly by addressing the vanishing gradient problem. Originally proposed by He et al. \cite{He15, He16}, ResNets simplify the highway network concept introduced by Srivastava et al. \cite{Srivastava15a, Srivastava15}. The key innovation of ResNets lies in their skip connections, which create direct pathways between layers \cite{Li18}. These connections improve gradient flow, making it possible to train much deeper networks. Veit et al. \cite{Veit16} conceptualized ResNets as an ensemble of paths of varying lengths, which effectively counteracts the vanishing gradient problem. Jastrzebski et al. \cite{JastrzSebski17} further explored the role of residual connections, showing how they facilitate iterative feature refinement and reduce overfitting. Additionally, Li et al. \cite{Li18} found that incorporating residual terms (skip connections) into plain networks smooths the loss function, leading to smaller backpropagation gradients and enhanced accuracy.

In the field of engineering, Lu et al. \cite{LuLu20} applied advanced neural networks using a multifidelity approach to extract material properties from instrumented indentation data. Additionally, Wang et al. \cite{Wang21} developed an enhanced fully-connected neural network architecture that incorporates transformer networks to improve predictive accuracy. In our latest work \cite{Amir23}, we introduce a residual-based method, called the square residual network (Sqr-ResNet), for solving function approximation and inverse PDEs. Here, we use this algorithm in addition to a simple highway network (Simple HwNet) to solve various types of PDEs and demonstrate the effect of this architecture on the loss of PDEs as well.

To leverage these advancements, numerous packages have been developed by various authors to facilitate solving PDE problems \cite{McClenny21, LuLu20d, Hennigh20, Koryagin19, Chen20}. One of the most well-known packages is DeepXDE \cite{LuLu20d}, which incorporates many of these strategies to enhance stability and performance.
In this paper, we introduce the Simple Highway Network and Squared ResNet, novel architectures designed to enhance neural network performance in solving multidimensional PDEs. These networks are straightforward to set up and significantly stabilize weight updating, thereby increasing stability in PINN. Our objectives include introducing these innovative architectures for improving accuracy and faster convergence, and demonstrating their superiority over plain neural networks. We conduct extensive experiments on various PDEs to illustrate the benefits of deeper architectures. To ensure a fair comparison, we utilize the DeepXDE package \cite{LuLu20d} for solving examples using plain networks and extend this package to integrate our proposed highway network-based architecture. Through rigorous analysis and comparisons, we highlight the advantages of the Squared ResNet, emphasizing its enhanced accuracy and stability.

The structure of the paper is organized as follows: Section 2 provides a review of plain neural networks, including both forward and backward operation networks. Section 3 discusses physics-informed neural networks for solving PDEs. Section 4 explores advancements in neural network architectures within the deep learning field. Section 5 details the experimental setup, presents the evaluation of results, and offers a discussion of the findings. Finally, Section 6 concludes the paper with a summary of contributions and suggestions for future research directions.

\section{Plain Neural Networks}
\subsection{Multi-Layer Perceptrons (MLPs)}

This section delves into the complex design and functional mechanisms of Multi-Layer Perceptrons (MLPs), which are fundamental to many deep learning systems. Denoted by $\mathcal{N}$, MLPs are precisely engineered to approximate a function $\textbf{u} : \textbf{q} \in \mathbb{R}^d \rightarrow \textbf{r} \in \mathbb{R}^D$ by systematically organizing artificial neurons across multiple layers.

\subsection{Layer Configuration}

The structure of a standard MLP includes several layers. The input layer, also known as the source layer, is crucial as it introduces the input data (dimension \(d\)) to the network. Following this, the hidden layers are positioned between the input and output layers. These hidden layers are responsible for executing complex computations, allowing the network to identify and understand intricate patterns and relationships in the data. Finally, the output layer, situated at the end, produces the network's predictions, often having a dimension \(D\).

The number of neurons in each layer is determined by the width of the layer, represented as \(p^{(k)}\). For a network with \(K\) hidden layers, the output vector of the \(k\)-th layer, represented as \(\textbf{q}^{(k)} \in \mathbb{R}^{p^{(k)}}\), becomes the input for the next layer. The signal that the input layer provides is denoted as \(\textbf{q}^{(0)} = \textbf{q} \in \mathbb{R}^d\).

Within each layer \(k\) (where \(1 \leq k \leq K + 1\)), the \(i\)-th neuron executes an affine transformation followed by a non-linear function. This process is described by the equation:

\begin{equation}
    h^{(k)}_i = W^{(k)}_{ij} {\rm{q}}^{(k-1)}_j + b^{(k)}_i,
\end{equation}
where \(1 \leq i \leq p^{(k)}\) and \(1 \leq j \leq p^{(k-1)}\). Subsequently, the output of the \(i\)-th neuron in layer \(k\) is obtained using the activation function:
\begin{equation}
    {\rm{q}}^{(k)}_i = \sigma(h^{(k)}_i),
\end{equation}
for \(1 \leq i \leq p^{(k)}\). In this context, \(W^{(k)}_{ij}\) and \(b^{(k)}_i\) denote the weights and biases of the \(i\)-th neuron in layer \(k\), respectively, and \(\sigma(\cdot)\) is the activation function, specifically \(\tanh\) in this instance. The neural network's overall function, \(\mathcal{N} : \mathbb{R}^d \rightarrow \mathbb{R}^D\), can be perceived as a sequence of alternating affine transformations and non-linear activations, as illustrated by the equations above. 
%

\subsection{Network Parameterization}

The network's parameters include all associated weights and biases, defined as \( \omega = \{\textbf{W}^{(k)}, \textbf{b}^{(k)}\}_{k=1}^{K+1} \). Each layer \( k \) is characterized by its weight matrix, denoted as \(\textbf{W}^{(k)}\), along with a bias vector, indicated as \(\textbf{b}^{(k)}\). 
Thus, the network $\mathcal{N}(\textbf{q}; \omega)$ encompasses a variety of parameterized functions, necessitating careful selection of \(\omega\) to ensure the network accurately approximates the desired function \(\textbf{u}(\textbf{q})\) at the input \(\textbf{q}\).

\subsection{Back-propagation for Gradient Computation}

In this section, we delve into the mechanics of gradient calculation during neural network training, with a particular focus on the back-propagation technique. For further insights, interested readers can refer to \cite{Schmidhuber15}. A crucial aspect of the training process involves computing gradients as the network advances.

Let us revisit the computation of the output $\textbf{q}^{(k+1)}$ at the $(k + 1)$-th layer:

\textbf{Affine Transformation:}
\begin{equation}\label{zeq_new}
    h^{(k+1)}_i = W^{(k+1)}_{ij} {\rm q}^{(k)}_j + b^{(k+1)}_i, \quad 1 \leq i \leq p^{(k+1)}, \quad 1 \leq j \leq p^{(k)}.
\end{equation}

\textbf{Non-linear Transformation:}
\begin{equation}\label{sigmaeq_new}
    {\rm q}^{(k+1)}_i = \sigma\left(h^{(k+1)}_i\right), \quad 1 \leq i \leq p^{(k+1)}.
\end{equation}

Considering a training instance denoted as $(\textbf{q}, \textbf{r})$, where $\textbf{q}^{(0)} = \textbf{q}$, the loss function can be computed through a forward pass:

For $k = 1, \ldots, K + 1$:
\begin{equation}\label{forward_new}
    \textbf{h}^{(k)} = \textbf{W}^{(k)} \textbf{q}^{(k-1)} + \textbf{b}^{(k)},
\end{equation}
\begin{equation}\label{activation_new}
    \textbf{q}^{(k)} = \sigma(\textbf{h}^{(k)}).
\end{equation}

\begin{figure}[!h]
\centering
\includegraphics[width=6.3in]{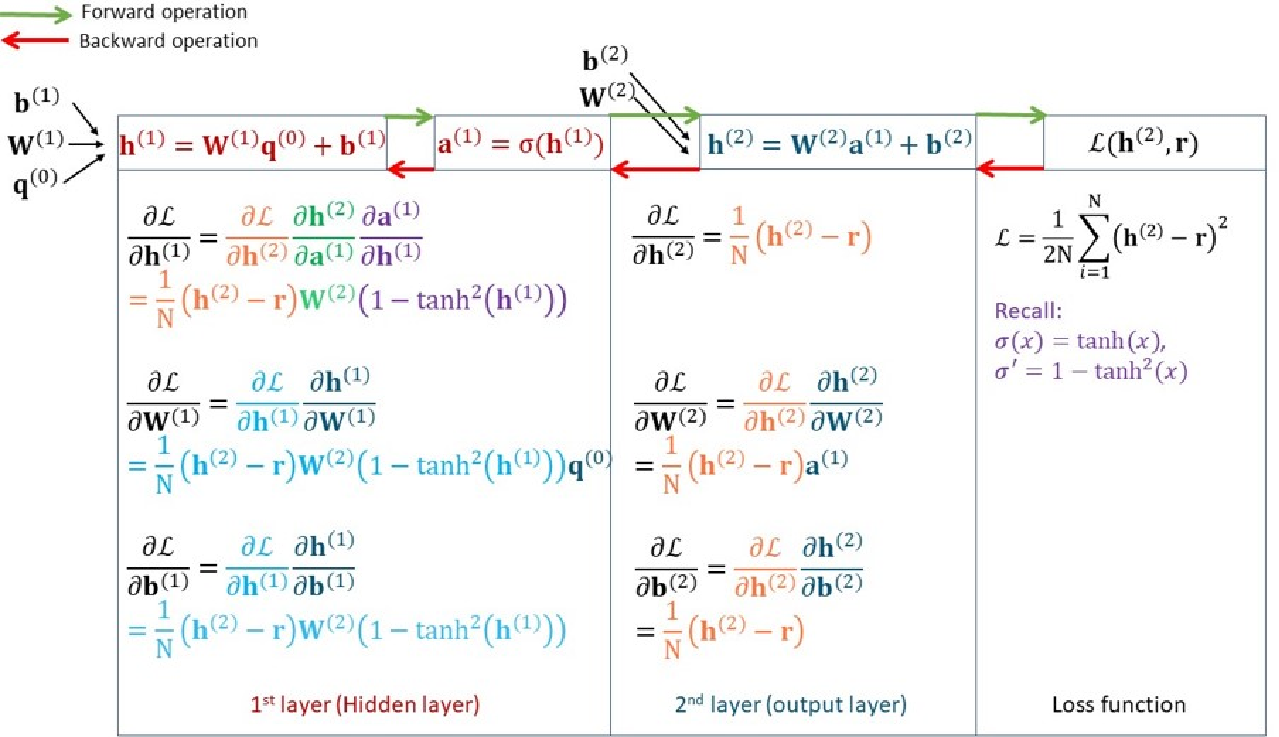}
\caption{Backpropagation expressions for a 2-layer network.} 
\label{backpropagation}
\end{figure}

The loss function is evaluated as:
\begin{equation}\label{lossfunc_new}
    {\mathcal{L}}(\omega) = \|\textbf{r} - {\mathcal{N}}({\textbf{q}}; \omega, \xi)\|^2,
\end{equation}
where  \( \xi \) denotes a fixed set of hyperparameters such as optimization method, network's length and width, etc. 
To update the network parameters, we need the derivatives $\frac{\partial {\mathcal{L}}}{\partial \omega}$, specifically $\frac{\partial {\mathcal{L}}}{\partial \textbf{W}^{(k)}}, \frac{\partial {\mathcal{L}}}{\partial \textbf{b}^{(k)}}$ for $1 \leq k \leq K + 1$. These derivatives are obtained by first deriving expressions for $\frac{\partial {\mathcal{L}}}{\partial {\textbf{h}}^{(k)}}$ and $\frac{\partial {\mathcal{L}}}{\partial \textbf{q}^{(k)}}$, then iteratively applying the chain rule:

\begin{equation}\label{chain_new}
    \frac{\partial {\mathcal{L}}}{\partial {\textbf{h}}^{(k)}} = \frac{\partial {\mathcal{L}}}{\partial \textbf{q}^{(K+1)}} \cdot \frac{\partial \textbf{q}^{(K+1)}}{\partial {\textbf{h}}^{(K+1)}} \cdot \frac{\partial {\textbf{h}}^{(K+1)}}{\partial \textbf{q}^{(K)}} \cdots \frac{\partial \textbf{q}^{(k+1)}}{\partial {\textbf{h}}^{(k+1)}} \cdot \frac{\partial {\textbf{h}}^{(k+1)}}{\partial \textbf{q}^{(k)}} \cdot \frac{\partial \textbf{q}^{(k)}}{\partial {\textbf{h}}^{(k)}}.
\end{equation}
The computation involves evaluating the following terms:
\begin{equation}\label{dL_dph_new}
    \frac{\partial {\mathcal{L}}}{\partial \textbf{q}^{(K+1)}} = -2(\textbf{r} - \textbf{q}^{(K+1)})^T,
\end{equation}
\begin{equation}\label{dz_dph_new}
    \frac{\partial {\textbf{h}}^{(k+1)}}{\partial \textbf{q}^{(k)}} = \textbf{W}^{(k+1)},
\end{equation}
\begin{equation}\label{dp_dz_new}
    \frac{\partial \textbf{q}^{(k)}}{\partial {\textbf{h}}^{(k)}} = \textbf{T}^{(k)} \equiv \text{diag}[\sigma'( {\textbf{h}}^{(k)}_1 ), \ldots , \sigma'( {\textbf{h}}^{(k)}_{p^{(k)}} )].
\end{equation}
These terms lead to the expression:
\begin{equation}\label{dL_dzh_new}
    \frac{\partial {\mathcal{L}}}{\partial {\textbf{h}}^{(k)}} = \textbf{T}^{(k)} \textbf{W}^{(k+1)T} \textbf{T}^{(k+1)} \cdots \textbf{W}^{(K+1)T} \textbf{T}^{(K+1)} [-2(\textbf{r} - \textbf{q}^{(K+1)})].
\end{equation}
Finally, an explicit expression for $\frac{\partial {\mathcal{L}}}{\partial \textbf{W}^{(k)}}$ is derived as:
\begin{equation}\label{bp}
    \frac{\partial {\mathcal{L}}}{\partial \textbf{W}^{(k)}} = \frac{\partial {\mathcal{L}}}{\partial {\textbf{h}}^{(k)}} \cdot \frac{\partial {\textbf{h}}^{(k)}}{\partial \textbf{W}^{(k)}} = \frac{\partial {\mathcal{L}}}{\partial {\textbf{h}}^{(k)}} \otimes \textbf{q}^{(k-1)}.
\end{equation}
In equation \eqref{bp}, $[\textbf{q} \otimes \textbf{r}]_{ij} = {\rm q}_i {\rm r}_j$ denotes the outer product. Therefore, to evaluate \( \frac{\partial {\mathcal{L}}}{\partial \textbf{W}^{(k)}} \), both \( \textbf{q}^{(k-1)} \), evaluated during the forward phase, and \( \frac{\partial {\mathcal{L}}}{\partial {\textbf{h}}^{(k)}} \), evaluated during back-propagation, are required. The expression for backpropagation for a 2-layer network (including one hidden layer and one output layer) is presented in Figure~\ref{backpropagation}. Numerical experiments on \eqref{bp} for a specific epoch (iteration) number are presented in Example 1, Figure~\ref{Ex0_3}.

\subsection{Training and Testing of MLPs}

Within the domain of supervised learning, the progression through iterative training and validation phases plays a pivotal role in refining and scrutinizing the efficacy of neural networks. Let \( \mathcal{R} = \{(\textbf{q}_i, {\textbf{r}}_i) : 1 \leq i \leq n\} \) constitute a dataset comprising paired samples encapsulating a target function \( \textbf{u} : \textbf{q} \rightarrow \textbf{r} \). The principal objective revolves around approximating this function via the utilization of the neural network \( \mathcal{N}(\textbf{q}; \omega, \xi) \), where \( \omega \) denotes the network's parameters and \( \xi \) encompasses hyperparameters encompassing characteristics such as depth, width, and the nature of the activation function.

The optimization journey of the network comprises two cardinal phases:

1. \textbf{Training:} During this stage, the neural network undergoes training utilizing the training set \( \mathcal{R}_{\text{train}} \), consisting of \( n_{\text{train}} \) data points. The primary objective is to determine the optimal parameters \( \omega^* \) by minimizing the training loss function \( \mathcal{L}_{\text{train}}(\omega) \), which is formally expressed as:
\[
\omega^* = \arg \min_\omega \mathcal{L}_{\text{train}}(\omega)
\]
The training loss function \( \mathcal{L}_{\text{train}}(\omega) \) is defined as the average squared Euclidean distance between the predicted outputs and the actual outputs over the training set, formulated as:
\[
\mathcal{L}_{\text{train}}(\omega) = \frac{1}{n_{\text{train}}} \sum_{i=1}^{n_{\text{train}}} \| \mathbf{r}_i - \mathcal{N}(\mathbf{p}_i; \omega, {\xi}) \|_2^2. 
\]
Here, \( \xi \) denotes a fixed set of hyperparameters. The optimal \( \omega^* \) is determined using an appropriate gradient-based optimization algorithm, as elaborated in Section 5. The mean-squared loss function \( \mathcal{L}_{\text{train}} \) serves as the primary criterion for the training process.

2. \textbf{Validation or Testing Phase:} Following the determination of the ``optimal'' network parameters and hyperparareters denoted by \( \omega^* \) and \( \xi^* \), respectively, it proceeds to undergo assessment utilizing the test set \( \mathcal{R}_{\text{test}} \). This assessment is conducted to gauge its efficacy on previously unobserved data, thereby confirming its performance beyond the initial training phases.

\section{Overview of Physics-Informed Neural Networks}

Physics-informed neural networks are used to solve nonlinear partial differential equations by inferring continuous solution functions \( \mathbf{u}(\mathbf{x}, {t}) \), where \(\mathbf{q} = (\mathbf{x}, {t})\), based on given physical equations. Consider a general nonlinear PDE of the form:
\begin{equation}\label{pde}
\begin{aligned}
\mathbf{u}_{t} + N_\mathbf{x}[\mathbf{u}] &= 0, & \mathbf{x} \in \Omega, & {t} \in [0, \mathcal{T}], \\
\mathbf{u}(\mathbf{x}, 0) &= I(\mathbf{x}), & \mathbf{x} \in \Omega, \\
\mathbf{u}(\mathbf{x}, t) &= B(\mathbf{x}, {t}), & \mathbf{x} \in \partial \Omega, & {t} \in [0, \mathcal{T}].
\end{aligned}
\end{equation}
We define \( \mathbf{x} = (x_1, x_2, \ldots, x_d) \in \mathbb{R}^d \) and \( {t} \) as the spatial and temporal coordinates, respectively, \( \mathbf{u}_{t} \) as the time derivative, and \( N_\mathbf{x}[\cdot] \) as a differential operator. The goal is to determine the solution function \( \mathbf{u}(\mathbf{x}, {t}) \) under these conditions.

Based on the foundational work of PINNs, a neural network \(  \mathcal{N}(\mathbf{x}, {t}; \omega, {\xi}) \) is used to approximate the solution function \( \mathbf{u}(\mathbf{x}, {t}) \).
Substituting \(  \mathcal{N}(\mathbf{x}, {t}; \omega, {\xi}) \) into the PDE \eqref{pde}, we define the residual:
\begin{equation}
r(\mathbf{x}, {t}; \omega, {\xi}) := \frac{\partial}{\partial {t}} \mathcal{N}(\mathbf{x}, {t}; \omega, {\xi}) + N_\mathbf{x}[\mathcal{N}(\mathbf{x}, {t}; \omega, {\xi})],
\end{equation}
where partial derivatives are computed using automatic differentiation. The loss function for PINNs is composed of multiple terms:
\begin{equation}
\mathcal{L}(\omega, {\xi}) =  \mathcal{L}_i(\omega, {\xi}) + \mathcal{L}_b(\omega, {\xi}) + \mathcal{L}_r(\omega, {\xi}),
\end{equation}
where \( \lambda_i \), \( \lambda_b \), and \( \lambda_r \) are weights for the initial condition loss \( L_i(\theta) \), boundary condition loss \( L_b(\theta) \), and residual loss \( L_r(\theta) \), respectively. These terms are defined as:
\begin{equation}
\begin{aligned}
\mathcal{L}_i(\omega, {\xi}) &= \frac{1}{N_i} \sum_{j=1}^{N_i} \left( \mathcal{N}(\mathbf{x}_j^i, 0; \omega, {\xi}) - I(\mathbf{x}_j^i) \right)^2, \\
\mathcal{L}_b(\omega, {\xi}) &= \frac{1}{N_b} \sum_{j=1}^{N_b} \left( \mathcal{N}(\mathbf{x}_j^b, {t}_j^b; \omega, {\xi}) - B(\mathbf{x}_j^b, {t}_j^b) \right)^2, \\
\mathcal{L}_r(\omega, {\xi}) &= \frac{1}{N_r} \sum_{j=1}^{N_r} r(\mathbf{x}_j^r, {t}_j^r; \omega, {\xi})^2,
\end{aligned}
\end{equation}
where \( \{\mathbf{x}_j^i, I(\mathbf{x}_j^i)\}_{j=1}^{N_i} \) are initial condition data points, \( \{(\mathbf{x}_j^b, {t}_j^b), B(\mathbf{x}_j^b, {t}_j^b)\}_{j=1}^{N_b} \) are boundary condition data points, and \( \{\mathbf{x}_j^r, {t}_j^r\}_{j=1}^{N_r} \) are internal collocation points. These points ensure the neural network satisfies initial and boundary conditions and minimizes the residuals.
Training involves optimizing the neural network parameters using gradient descent methods such as Adam, SGD, or L-BFGS. By minimizing \( \mathcal{L}(\omega, {\xi}) \), the neural network \( \mathcal{N}(\mathbf{x}, {t}; \omega, {\xi}) \) approximates the solution \( \mathbf{u}(\mathbf{x}, {t}) \) for the PDE.

\section{Advanced Deep Neural Network Architectures}
\subsection{Highway Networks}
\begin{figure}[!h]
\centering
\includegraphics[width=6.3in]{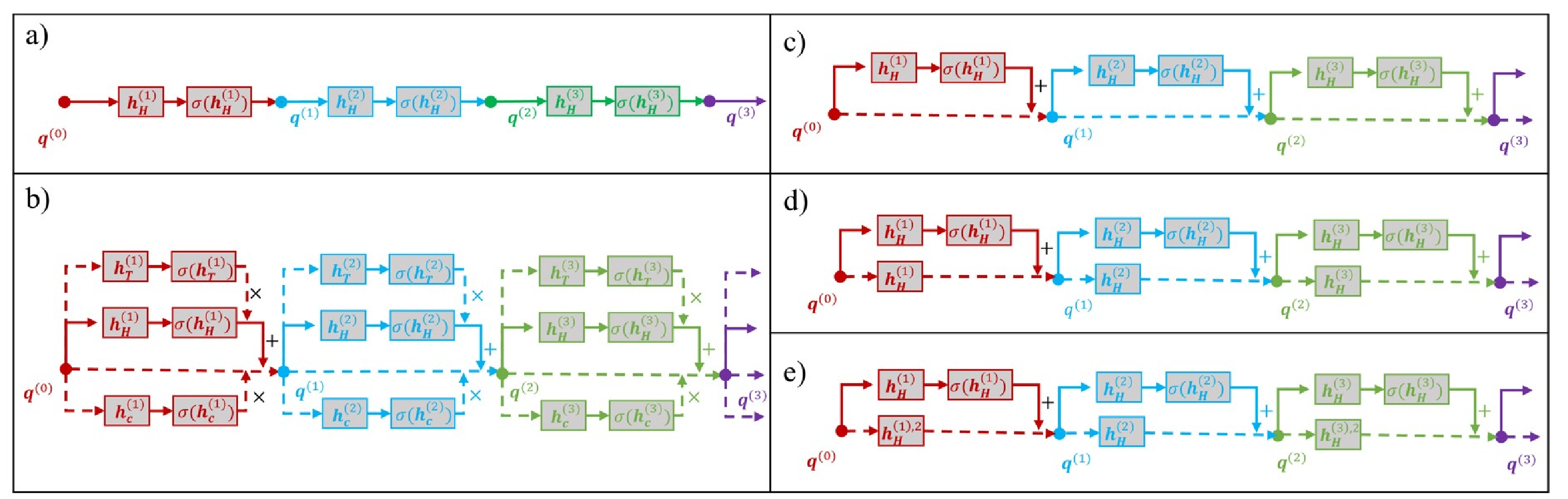}
\caption{four neural network architectures: (a) a basic neural network (Plain Net), (b)  Highway network (Hw), and (c) ResNet (d) the proposed simple highway network (simple HwNet) (e) Square-Residual network (Sqr-ResNet).} 
\label{ExNet}
\end{figure}

In this subsection, we discuss Highway Networks, a variant of deep neural networks specifically designed to improve the transmission of information through multiple layers \cite{Srivastava15a,Srivastava15}. These networks employ gating mechanisms that regulate the flow of information, thus facilitating selective processing and retention of the input data. A standard neural network can be depicted as (Figure~\ref{ExNet}(a)):
\begin{equation} \label{qk_rule}
\textbf{q}^{(k)} = H^{(k)}(\textbf{q}^{(k-1)}, \textbf{W}_{f}^{(k)}).
\end{equation}
In \eqref{qk_rule}, \( \textbf{q} \) denotes the input to a particular layer, and \( H^{(k)}(\textbf{q}^{(k-1)}, \textbf{W}_{f}^{(k)}) \) represents the operations performed at the \( k \)-th layer.
Highway Networks (Figure~\ref{ExNet}(b)) employ two distinct gating mechanisms. The first mechanism, known as the transform gate \( T \), is responsible for managing nonlinear transformations. The second mechanism, the carry gate \( C \), controls the flow of activation from one layer to the next. These mechanisms are mathematically represented as follows:
\begin{equation}\label{hn}
\textbf{q}^{(k)} = H^{(k)}(\textbf{q}^{(k-1)}, \textbf{W}_{f}^{(k)}) \cdot T^{(k)}(\textbf{q}^{(k-1)}, \textbf{W}_{T}^{(k)}) + \textbf{q}^{(k-1)} \cdot C^{(k)}(\textbf{q}^{(k-1)}, \textbf{W}_{C}^{(k)}).
\end{equation} \noindent
In \eqref{hn}, $H^{(k)}(\textbf{q}^{(k-1)}, \textbf{W}_{f}^{(k)})$ denotes the combination of linear and nonlinear functions performed by the highway neural network at layer \( k \).
 This design allows Highway Networks to simultaneously learn both direct and skip connections, which aids in the efficient training of very deep networks by maintaining essential information throughout the layers. Highway Networks have shown promising results in various applications such as speech recognition, natural language processing, and image classification, demonstrating their utility in deep neural network training \cite{Srivastava15a,Srivastava15}. However, additional training is needed for the carry and transfer gates. Equation \eqref{hn} requires that the dimensions of $\mathbf{p}^{(k)}$, $T^{(k)}(\mathbf{p}^{(k-1)}, \mathbf{W}_{T}^{(k)})$, and $C^{(k)}(\mathbf{p}^{(k-1)}, \mathbf{W}_{C}^{(k)})$ stay the same.
The desired outcome can be achieved through several methods. One method involves substituting $\mathbf{p}^{(k)}$ with $\hat{\mathbf{p}}^{(k)}$, which is derived by either sub-sampling or zero-padding $\mathbf{p}^{(k)}$ as needed. Another method is to use a basic layer without highways to modify the dimensionality before proceeding with the addition of highway layers. In the networks presented below, both the challenges of additional training and dimensionality are effectively resolved.

\subsection{Residual Networks}

Residual Networks (Figure~\ref{ExNet}(c)) are a streamlined adaptation of Highway Networks \cite{Greff17}, where the transformation is represented by the addition of the input and a residual component. Known as ResNets \cite{He15, He16}, these architectures have gained significant attention in the neural network community. They feature residual modules, labeled as \( H^{(k)} \), and include skip connections that bypass these modules, facilitating the construction of very deep networks. This design enables the formation of residual blocks, which are groups of layers within the network.
The output \(\textbf{q}^{(k)}\) at the \(k\)-th layer is determined as follows:
\begin{equation}
\textbf{q}^{(k)} = H^{(k)}(\textbf{q}^{(k-1)}, \textbf{W}_{H}^{(k)}) + \textbf{q}^{(k-1)}. \label{eq:residual}
\end{equation}
Here, the function \( H^{(k)}(\textbf{q}^{(k-1)}, \textbf{W}_{H}^{(k)}) \) represents a series of operations including linear transformations and activation functions at the \(k-1\) layer. This structure is characteristic of residual networks for \( 1 \leq k \leq K \). When comparing the residual network described by \eqref{eq:residual} with the highway network defined by \eqref{hn}, it becomes apparent that both the transfer gate \( (T) \) and carry gate \( (C) \) are set to 1 \cite{Greff17}, as shown below:
\begin{equation}
T^{(k)}(\textbf{q}^{(k-1)}, \textbf{W}_{T}^{(k)}) = 1,
\end{equation}
and
\begin{equation}
\textbf{q}^{(k-1)} \cdot C^{(k)}(\textbf{q}^{(k-1)}, \textbf{W}_{C}^{(k)}) = \textbf{q}^{(k-1)}.
\end{equation}
This indicates that no further modifications are required.

\subsection{A Simple Highway Network (Simple HwNet)}
Here, we present a streamlined version of the highway network, illustrated in Figure~\ref{ExNet}(d):
\begin{equation}
\textbf{q}^{(k)} = H^{(k)}(\textbf{q}^{(k-1)}, \textbf{W}_{f}^{(k)})+ \textbf{h}^{(k)}.  \label{eq:simplehw}
\end{equation}
In this formulation, $\textbf{h}^{(k)} = \textbf{W}_{H}^{(k)} {\textbf{q}}^{(k-1)} + \textbf{b}^{(k)}$. When comparing this simplified highway network \eqref{eq:simplehw} with the original structure in \eqref{hn}, we note that in the simplified model, we define 
\begin{equation}
T^{(k)}(\textbf{q}^{(k-1)}, \textbf{W}_{T}^{(k)}) = 1,
\end{equation}
and 
\begin{equation}
\textbf{q}^{(k-1)}\cdot C^{(k)}(\textbf{q}^{(k-1)}, \textbf{W}_{C}^{(k)}) = \textbf{W}^{(k)}_{H} {\textbf{q}}^{(k-1)} + \textbf{b}^{(k)}.
\end{equation}
This indicates that the weight parameters for both $\mathbf{W}^{(k)}_{H}$ and $\mathbf{W}^{(k)}_{C}$ in the simplified highway network remain the same, thus removing the need for further training compared to the original highway network. Moreover, this approach obviates the requirement for matrix size matching since the weights are considered within the basic network framework.

\subsection{Proposed Squared Residual Network (Sqr-ResNet)}
In this work, we introduce a straightforward highway network. The equation \eqref{hn} can be reformulated as illustrated in Figure~\ref{ExNet}(e):
\begin{equation}
\textbf{q}^{(k)} = H^{(k)}(\textbf{q}^{(k-1)}, \textbf{W}_{H}^{(k)}) + \textbf{h}^{(k)} \odot \textbf{h}^{(k)}, \label{eq:sqrhw}
\end{equation}
where $\odot$ signifies element-wise multiplication. Here, $\textbf{h}^{(k)}$ is given by $\textbf{W}_{H}^{(k)} \textbf{q}^{(k-1)} + \textbf{b}^{(k)}$. 

To evaluate the proposed SqrHw network, represented by \eqref{eq:sqrhw}, against the original model in \eqref{hn}, we observe that in the simplified model, we assume
\begin{equation}
T^{(k)}(\textbf{q}^{(k-1)}, \textbf{W}_{T}^{(k)}) = 1,
\end{equation} 
and the term
\begin{equation}\label{sqrhwcarry}
\textbf{q}^{(k-1)} \cdot C^{(k)}(\textbf{q}^{(k-1)}, \textbf{W}_{C}^{(k)}) = (\textbf{W}_{H}^{(k)} \textbf{q}^{(k-1)} + \textbf{b}^{(k)}) \odot (\textbf{W}_{H}^{(k)} \textbf{q}^{(k-1)} + \textbf{b}^{(k)}).
\end{equation}
In this context, the $\odot$ operation denotes element-wise multiplication, indicating that the carry gate performs a similar transformation on the input $\textbf{q}^{(k-1)}$ as seen in the original highway network, but introduces extra element-wise alterations. Equation \eqref{sqrhwcarry} implies that the weight updates remain consistent $(\textbf{W}^{(k)}_{H} = \textbf{W}^{(k)}_{C})$ with those in a standard neural network, thus eliminating the need for additional training or alignment of dimensions.




\section{Results and Discussions}\label{section:numexa}
In this study, we use the notations NP, NL, and NN to denote the number of data points (training), layers, and neurons in each layer, respectively. Throughout all subsequent examples, unless otherwise specified, we consider \(100^2\) validation data points. We compare three algorithms:

\begin{itemize}
  \item {\tt Plain Net (Plain Neural Network):} \\A plain neural network, also called a feedforward neural network or MLP.
  \item {\tt Simple HwNet (Simple Highway Network):} \\Residual terms are added to every other layer.
  \item {\tt Sqr-ResNet (Squared Residual Network):} \\The squared power of residual terms is added to every other layer.
\end{itemize}

\noindent
Our study shows that adding residual terms to every layer or every other layer does not change the results in terms of accuracy, but it improves efficiency regarding computational time. Therefore, we use every other layer for the residual terms.

The DeepXDE package \cite{deepxde21} is utilized for physics-informed neural network modeling with the following settings: \texttt{PyTorch} selected as the backend, and a random seed of {\tt 12345} is used for point generation. Computations are performed using {\tt float32} as the default in DeepXDE.

For optimization, the \texttt{L-BFGS} (Limited-memory Broyden-Fletcher-Goldfarb-Shanno) optimizer is employed with a maximum of {\tt 5000} iterations, a gradient tolerance of  \(1 \times 10^{-8}\), and a learning rate of  \(1 \times 10^{-3}\).
The neural network is configured with the activation function \texttt{tanh}, and the initializer is set to \texttt{Glorot uniform}. Each hidden layer has a fixed number of 50 neurons (NN), unless specified otherwise.
The DeepXDE's source code has been updated to incorporate ResNet and SqrResNet algorithms.

Tables in this section present train loss and test loss, indicating the mean square error $ \frac{1}{n} \sum_{i=1}^{n} \left( {u}_i - {N}_{i} \right)^2 $ over function interpolation and PDEs, respectively. The error and test metrics in figures are based on L2 norm error $\frac{\| {u} - {N} \|_2}{\| {u} \|_2}$, where  $u$ and $N$ denote exact and approximated solutions, respectively. \\
In tables, the term ``status'' denotes three situations as follows:
\begin{itemize}
  \item \texttt{completed:} Indicates that all predefined steps for the network have been successfully executed, or the network has converged even in a smaller number of pre-defined steps.
  \item \texttt{not trained:} Denotes a scenario where, from the beginning, a large loss was observed, and the situation further deteriorates as the steps progress.
  \item \texttt{diverged:} Signifies that the computations diverged after some initially successful iterations. Divergence can take two forms: (i) a significant increase in loss and error, or (ii) the occurrence of NaN due to SVD convergence issues.
\end{itemize}
\noindent
The numerical experiments were conducted on a computer with an Intel(R) Core(TM) i9-9900 CPU operating at 3.10 GHz and equipped with 64.0 GB of RAM.

Although the DeepXDE package is well-known and very user-friendly, we encountered some limitations when using it to solve our problems. Specifically, the package records the loss function output every 1000 epochs, and we were unable to adjust this frequency to a smaller interval. As a result, the plots generated display metric results (loss and errors) at every 1000 epochs. This issue has been reported to the author of the package as error \#1607. However, this limitation does not affect the final results and only makes the demonstration slightly less informative. Therefore, we continue to use the package to solve Examples 2-5 using both the plain network and the proposed networks fairly.

In the following section, five examples are presented (see Table~\ref{tab:ex0}). The first example demonstrates the importance of weight updating in a network's performance for function approximation problems, which is a preliminary stage in solving PDEs. We will show that the choice of architecture has a great effect on the weight updating. We also examine the gradient of the loss with respect to the updated weights to assess the quality of backpropagation. Subsequent examples involve solving various PDEs to illustrate the advantages of the proposed residual-based architectures in terms of stability and accuracy.

\begin{table}[!h]
  \captionof{table}{Study examples overview.}\label{tab:ex0}
  \begin{center}
    \begin{threeparttable}
      \begin{tabular}{|llll|} \hline
Example&Equation& Dimension & Time-dependent  \\ \hline
\ref{e1}&Function Approximation&2D&No\\\hline 
\ref{e4}&PDE: Diffusion-Reaction&1D&Yes\\\hline 
\ref{e6}&PDE: Heat&1D&Yes\\\hline 
\ref{e7}&PDE: Schrödinger&1D&Yes\\\hline 
\ref{e8}&PDE: Elastostatic&2D&No\\
\hline
      \end{tabular}
    \end{threeparttable}
  \end{center}
\end{table}

\begin{example}\label{e1}\rm{
We use an example of an approximation function to demonstrate the strength of the proposed architectures, both the simple highway network and the square ResNet. The function to be approximated (interpolated) is given by:

\begin{equation}
{\rm F} = \sin(3x_1) \cdot \cos(3x_2)
\end{equation}
where $x_1,x_2 \in [0,1]^2$.
Figure~\ref{Ex0_1} shows the validation error (L2 norm error) for approximating \( F \) using 500 collocation points, 10 hidden layers, and 50 neurons in each layer. From this figure, it is evident that the Plain Network is less accurate compared to both the proposed simple highway network and the square Residual network. Furthermore, the square ResNet demonstrates better accuracy with fewer epochs, converging at about 2700 epochs, whereas the simple highway network requires  about 3200 epochs to converge.

\begin{figure}
\centering%
\includegraphics[width=2.95in]{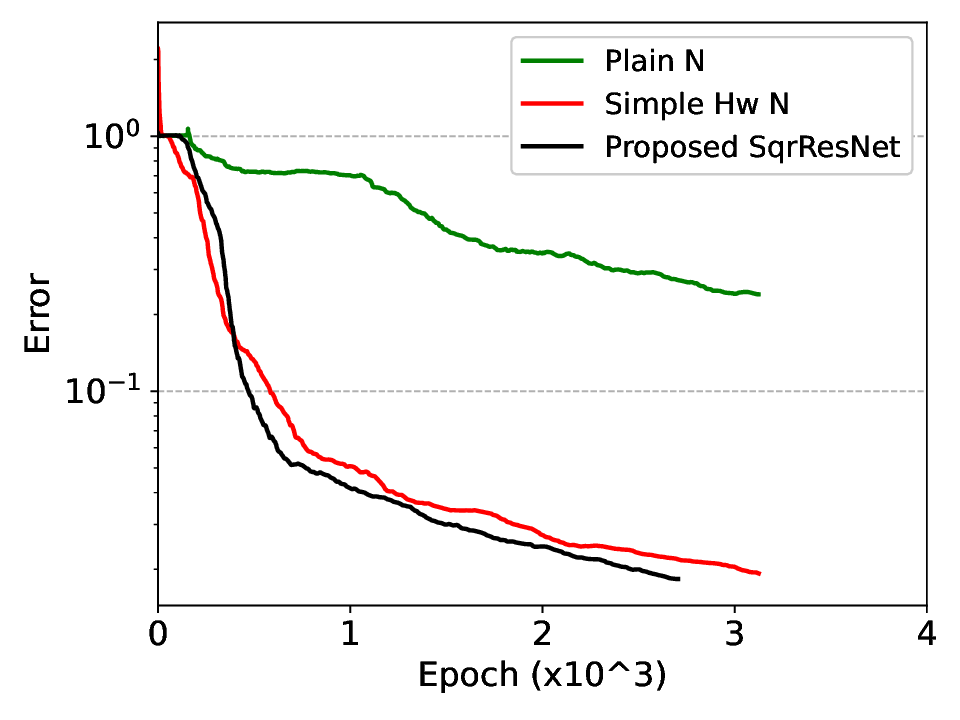}
\caption{Example 1: Validation error (L2 norm error) for three networks.} \label{Ex0_1}
\end{figure}

Figure~\ref{Ex0_2} displays the final approximated function in the top panel and the Frobenius norms of weights with respect to the epoch in the bottom panel. This figure highlights the relationship between accuracy and weight updating of the network. The top panel illustrates that both proposed networks outperform the Plain Network, which shows significant error near the boundary. In the bottom panel, a comparison of Figures~\ref{F11_weight_pn} and~\ref{F11_weight_shw} reveals that the weight norms for the Plain Network fluctuate significantly and increase more than those from the Simple Highway Network architecture. Moreover, the simple highway network in Figure~\ref{F11_weight_shw} shows a larger rise compared to the square-ResNet in Figure~\ref{F11_weight_sn_flat}. For the Sqr-ResNet case, the weights become stable after approximately 2500 epochs while the simple highway network requires slightly more epochs to stabilize the weight updating. A comparison between the error (Figure~\ref{Ex0_1}) and the weight updating in the bottom panel of Figure~\ref{Ex0_2} indicates that more stable weight updating leads to more accurate results and better convergence.

\begin{figure}
\centering%
\subfigure[Plain Net]{ \label{surf_pn}
\includegraphics[width=1.95in]{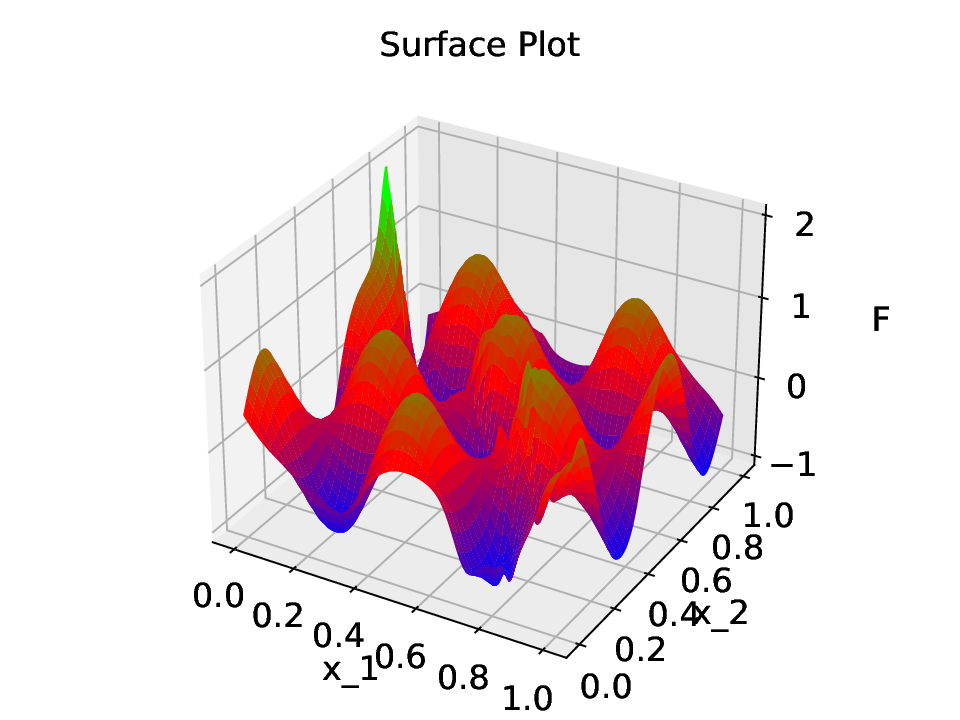}}
\subfigure[Simple HwNet]{ \label{surf_shw}
\includegraphics[width=1.95in]{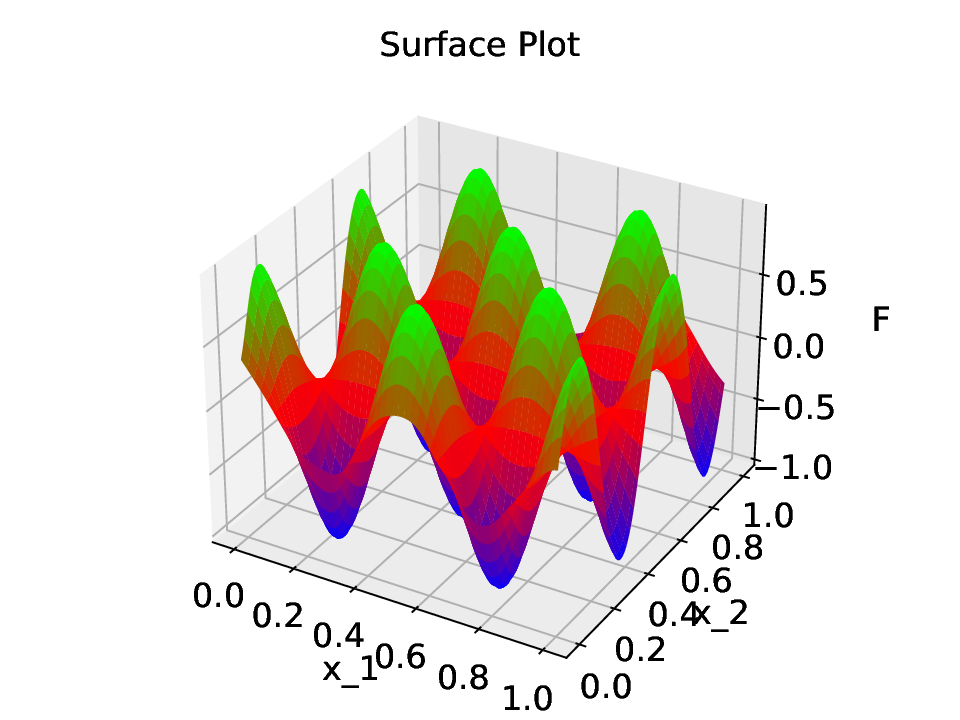}}
\subfigure[Sqr-ResNet]{ \label{surf_sn}
\includegraphics[width=1.95in]{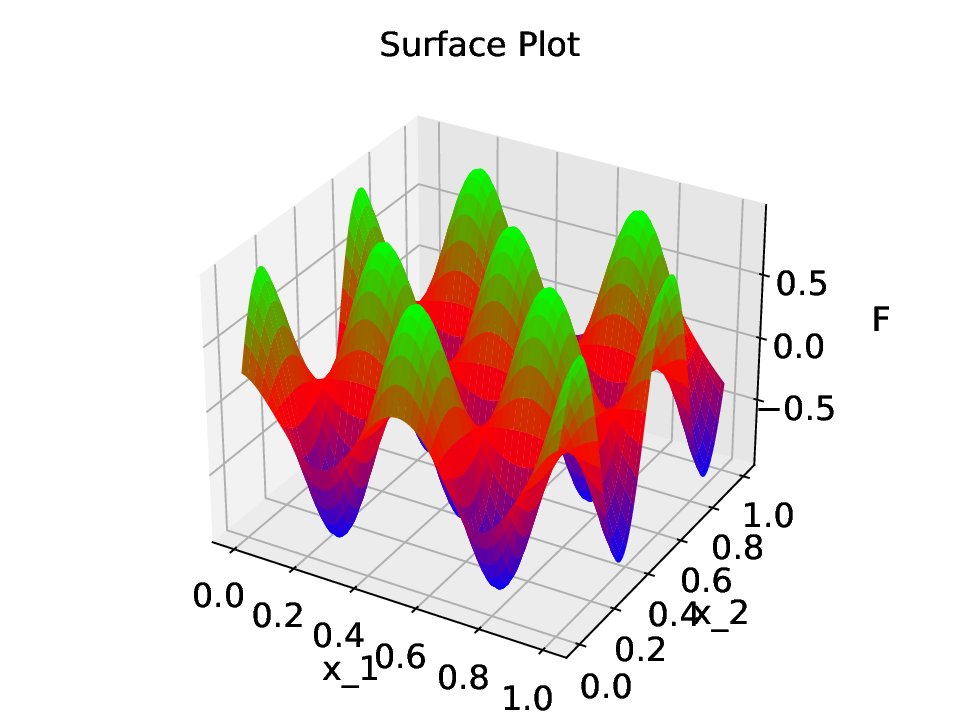}}
\subfigure[Plain Net]{ \label{F11_weight_pn}
\includegraphics[width=1.95in]{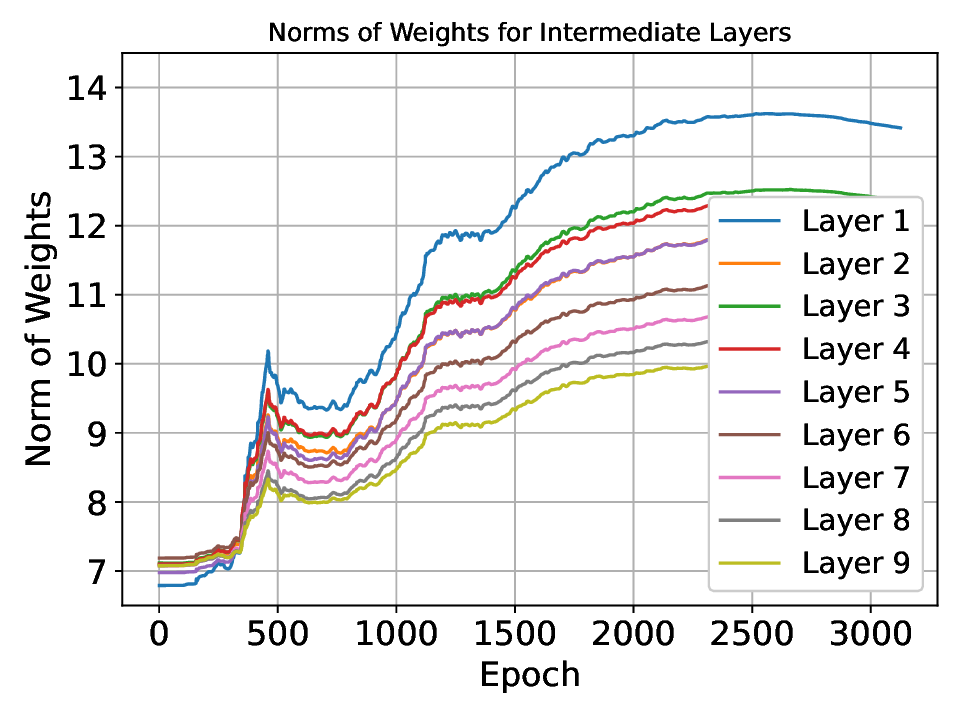}}
\subfigure[Simple HwNet]{ \label{F11_weight_shw}
\includegraphics[width=1.95in]{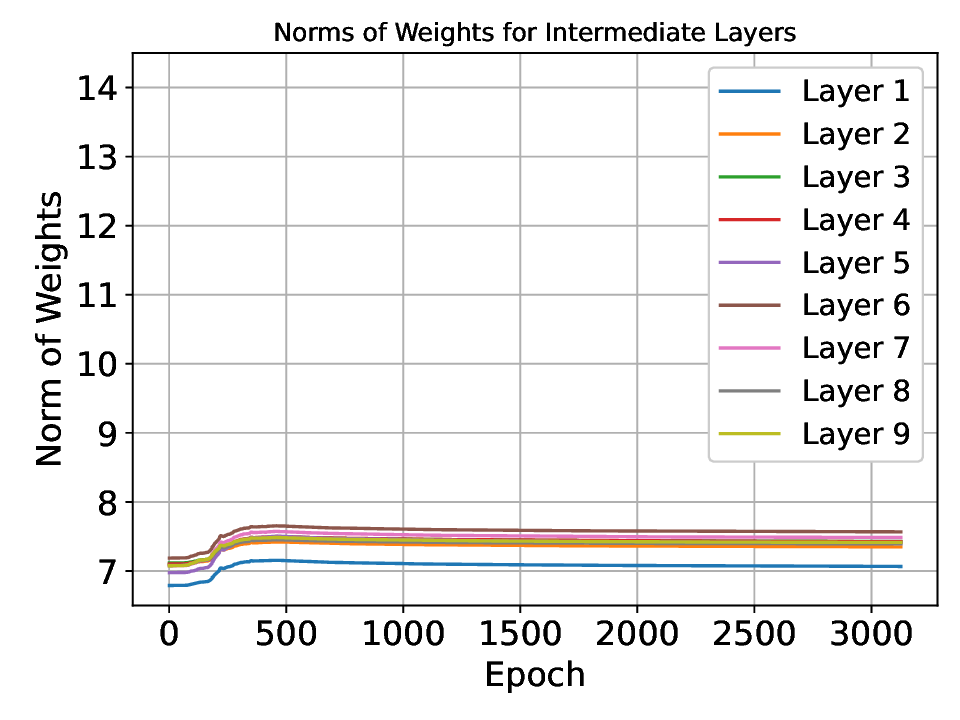}}
\subfigure[Sqr-ResNet]{ \label{F11_weight_sn_flat}
\includegraphics[width=1.95in]{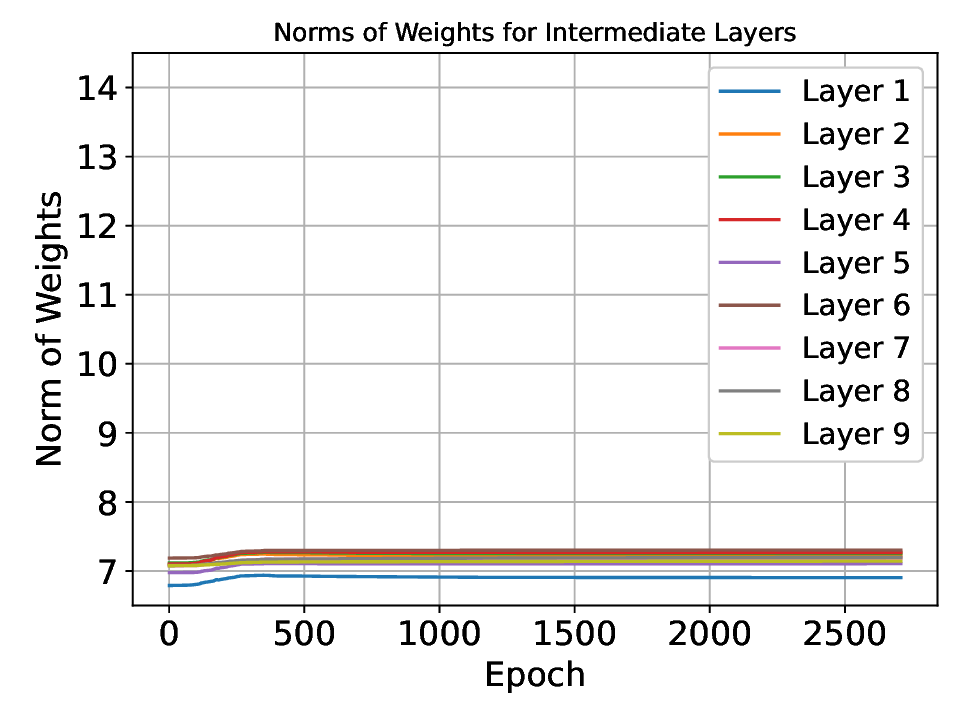}}
\caption{Example 1: Approximated function and norm of the error for three networks. Left panel: PINN;  middle panel: Simple highway network; right panel: square ResNet.} \label{Ex0_2}
\end{figure}

We plot the histogram of the norm of the gradient of loss with respect to weight at Epoch = 1000 for layer numbers 3 and 6 in Figure~\ref{Ex0_3}. To enhance visibility, we omit the results for the simple highway network, as they are similar to those of the square-ResNet architecture. As reported by Li et al. \cite{Li18}, residual data helps create a smoother loss function, leading to better convergence and smaller backpropagation gradients. 
 A ``good'' network architecture also helps the optimization method to be on the way to find the global minimal whereas plain network is prone to finish the optimization process in a local minimal. This can be recognized by the large error in results when using the plain network while  residual-based architectures can finish it  in a much smaller error. Our intense numerical tests in  this example and following examples show that the effect of the good network is more evident when the network includes a large number of layers. A large network indicates a larger number of computations which increases the risk of round-off error. Following these, Figure~\ref{Ex0_3} shows  Sqr-ResNet, compared to the Plain Network, results in much smaller gradient values with higher frequency.

\begin{figure}
\centering%
\subfigure[Layer=3]{ \label{histo}
\includegraphics[width=2.95in]{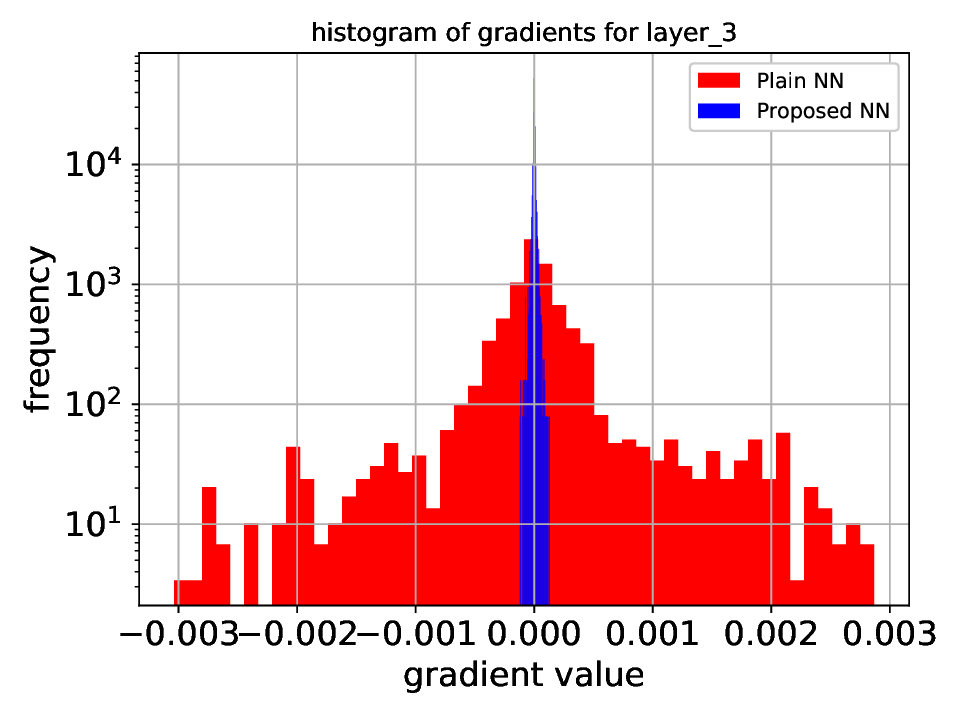}}
\subfigure[Layer=6]{ \label{histo6}
\includegraphics[width=2.95in]{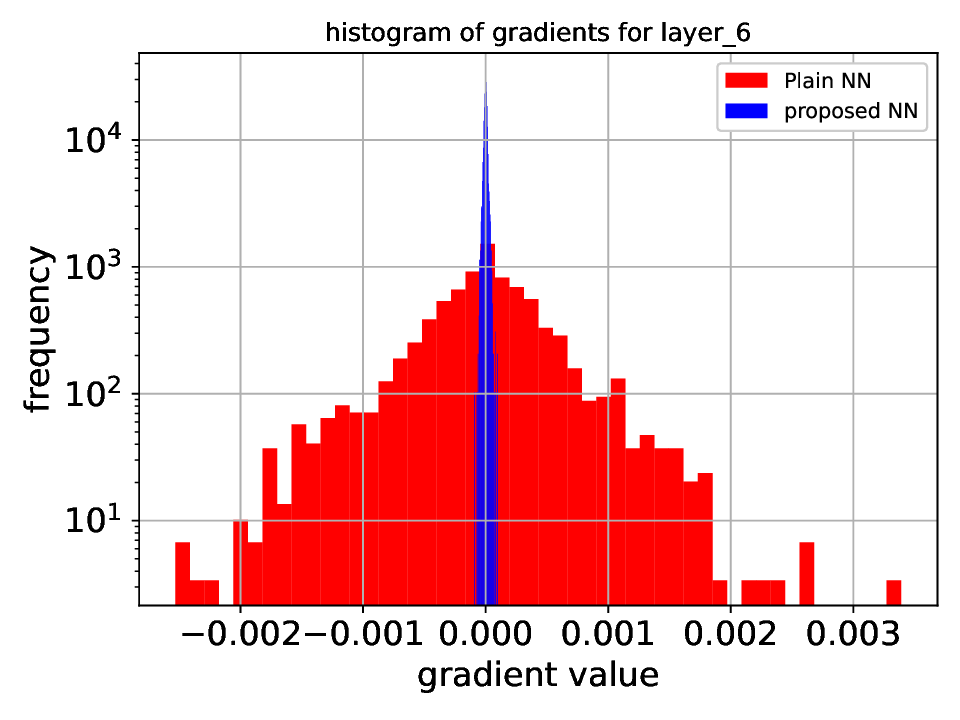}}
\caption{Example 1: The histogram of gradient of loss function with respect to the weights for the plain network and proposed square-ResNet.} \label{Ex0_3}
\end{figure}

Therefore, we conclude that this simple example demonstrates that an architecture incorporating residual terms can better approximate functions. This function serves as the input for the physics-informed neural network architecture. In the following examples, we show that the proposed residual-based architectures enhance the accuracy and stability of physics-informed neural networks.
}
\end{example}

\begin{example}\label{e4}\rm{
We aim to solve the 1D diffusion-reaction equation given by:

\begin{equation}
    \frac{\partial u}{\partial t} = d \frac{\partial^2 u}{\partial {x}_1^2} + e^{-t} \left( \frac{3}{2} \sin(2{x}_1) + \frac{8}{3} \sin(3{x}_1) + \frac{15}{4} \sin(4{x}_1) + \frac{63}{8} \sin(8{x}_1) \right)
\end{equation}
where \( d = 1 \).
subject to the initial condition at \( t=0 \):

\begin{equation}
    u({x}_1, 0) = \sin({x}_1) + \frac{1}{2}\sin(2{x}_1) + \frac{1}{3}\sin(3{x}_1) + \frac{1}{4}\sin(4{x}_1) + \frac{1}{8}\sin(8{x}_1), \quad {x}_1 \in [-\pi, \pi]
\end{equation}
and the Dirichlet boundary conditions:
\begin{equation}
    u(t, -\pi) = u(t, \pi) = 0, \quad t \in [0, 1].
\end{equation}
The exact solution to this equation is:
\begin{equation}
    u({x}_1, t) = e^{-t} \left( \sin({x}_1) + \frac{1}{2} \sin(2{x}_1) + \frac{1}{3} \sin(3{x}_1) + \frac{1}{4} \sin(4{x}_1) + \frac{1}{8} \sin(8{x}_1) \right).
\end{equation}

\begin{table}[!h]
  \captionof{table}{Example 2: Diffusion-reaction equation.}\label{tab:ex4}
  \begin{center}
   \resizebox{\textwidth}{!}{
    \begin{threeparttable}
      \begin{tabular}{|c|ccl|ccl|ccl|} \hline
  & \multicolumn{3}{c|}{Plain Net}  
  & \multicolumn{3}{c|}{Simple HwNet}
  & \multicolumn{3}{c|}{Sqr-ResNet}
   \\
         NL&train loss &error &status       
   &train loss &error &status
   &train loss &error &status  \\ \hline   
5    
&2.5e-4 &7.4e-4 &completed   
&3.5e-6  &9.4e-5 &completed   
&8.0e-6  &2.9e-4 &completed \\\hline 

10   
&9.8e-1 &3.9e-2 &diverged\tnote{*}   
&6.9e-6  &1.3e-4 &completed   
&8.5e-6  &1.9e-4 &completed  \\\hline 

15   
&2.6e-5 &3.0e-4 &completed   
&2.3e-6  &1.0e-4 &completed   
&\textbf{1.5e-6}  &\textbf{6.5e-5} &\textbf{completed} \\\hline 

20   &\ding{55} &\ding{55} &not trained 
&3.9e-6  &1.1e-4 &completed   
&2.8e-6  &1.0e-4 &completed \\ \hline

30   &\ding{55} &\ding{55} &not trained 
&\ding{55} &\ding{55} &not trained
&3.5e-6  &1.1e-4 &completed \\ 
\hline
      \end{tabular}
      \begin{tablenotes}
        \item[*]  Diverged after step = 1000.
\      \end{tablenotes}
    \end{threeparttable}}
  \end{center}
\end{table}

\begin{figure}
\centering%
\subfigure[loss, Plain Net]{ \label{Ex4_1DDiffRea_50x10_loss_p}
\includegraphics[width=1.95in]{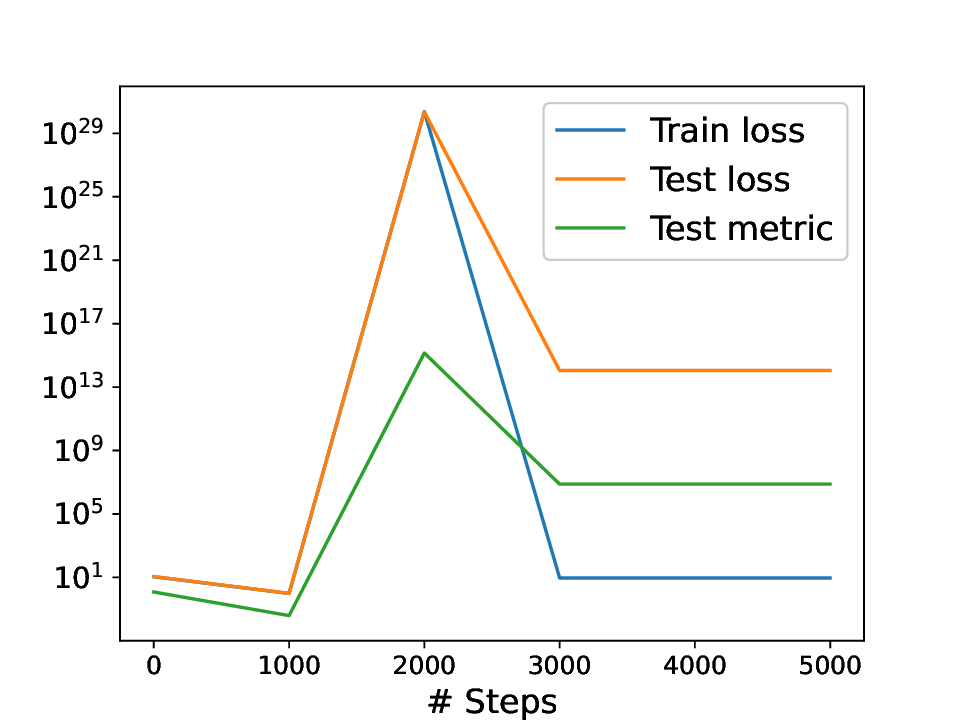}}
\subfigure[loss, Simple HwNet]{ \label{Ex4_1DDiffRea_50x10_loss_pr}
\includegraphics[width=1.95in]{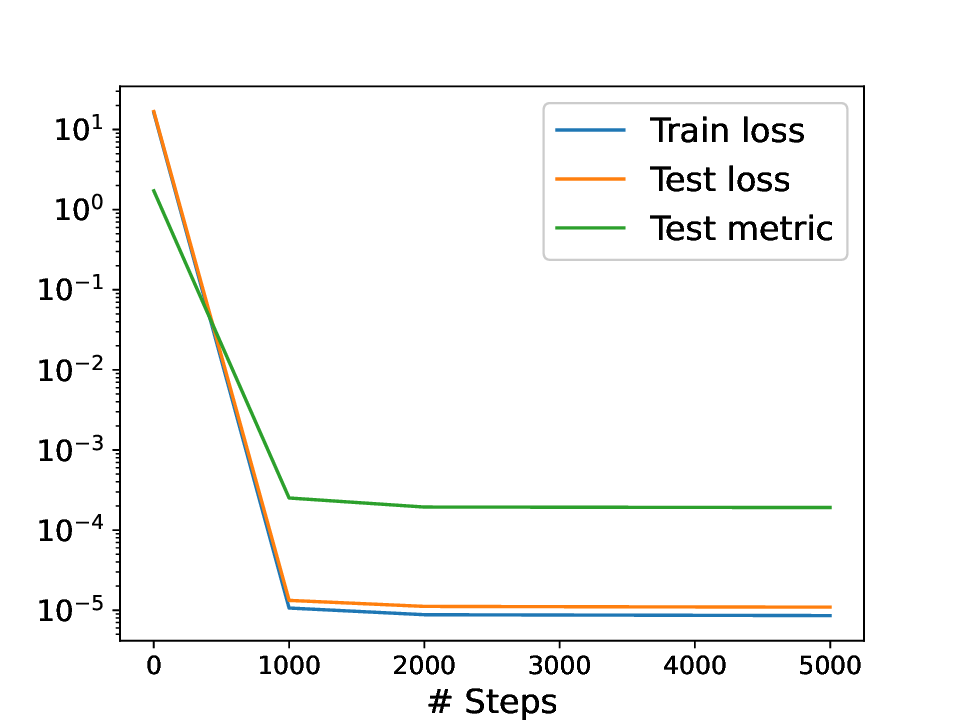}}
\subfigure[loss, Sqr-ResNet]{ \label{Ex4_1DDiffRea_50x10_loss_r}
\includegraphics[width=1.95in]{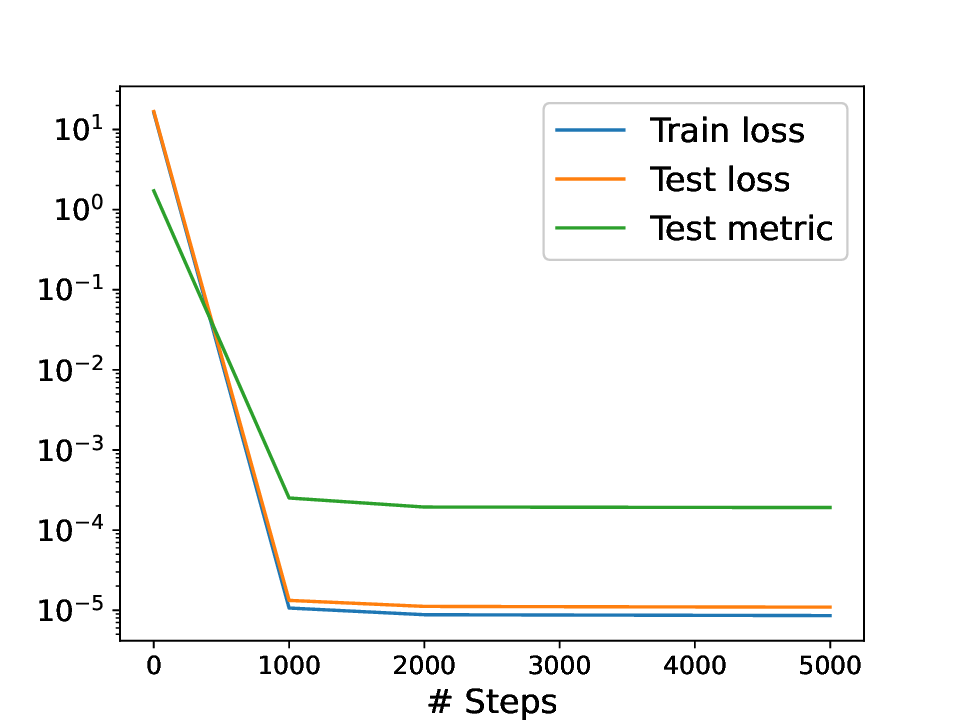}}
\subfigure[error, Plain Net]{ \label{Ex4_1DDiffRea_50x10_err_p}
\includegraphics[width=1.95in]{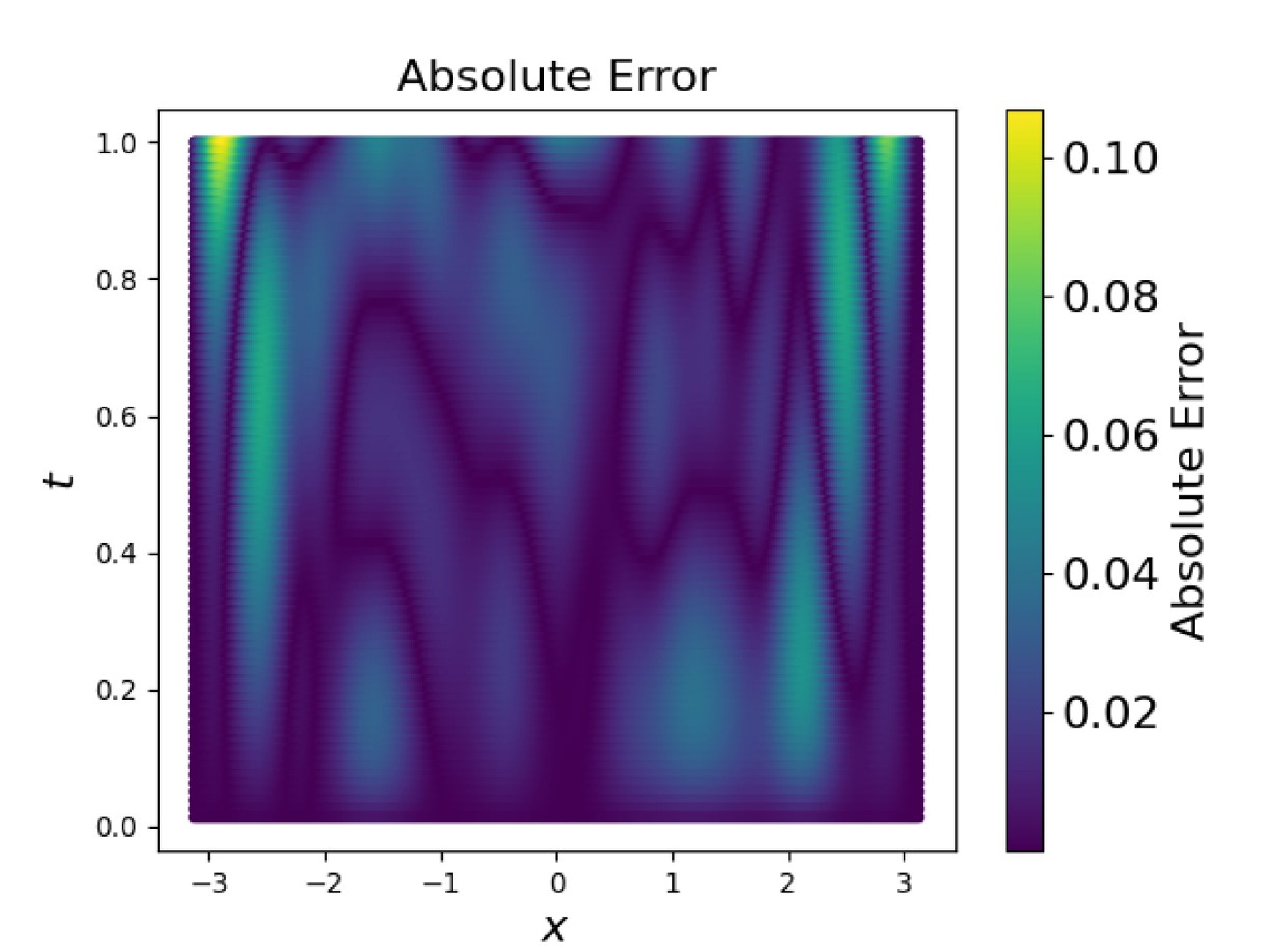}}
\subfigure[error, Simple HwNet]{ \label{Ex4_1DDiffRea_50x10_err_pr}
\includegraphics[width=1.95in]{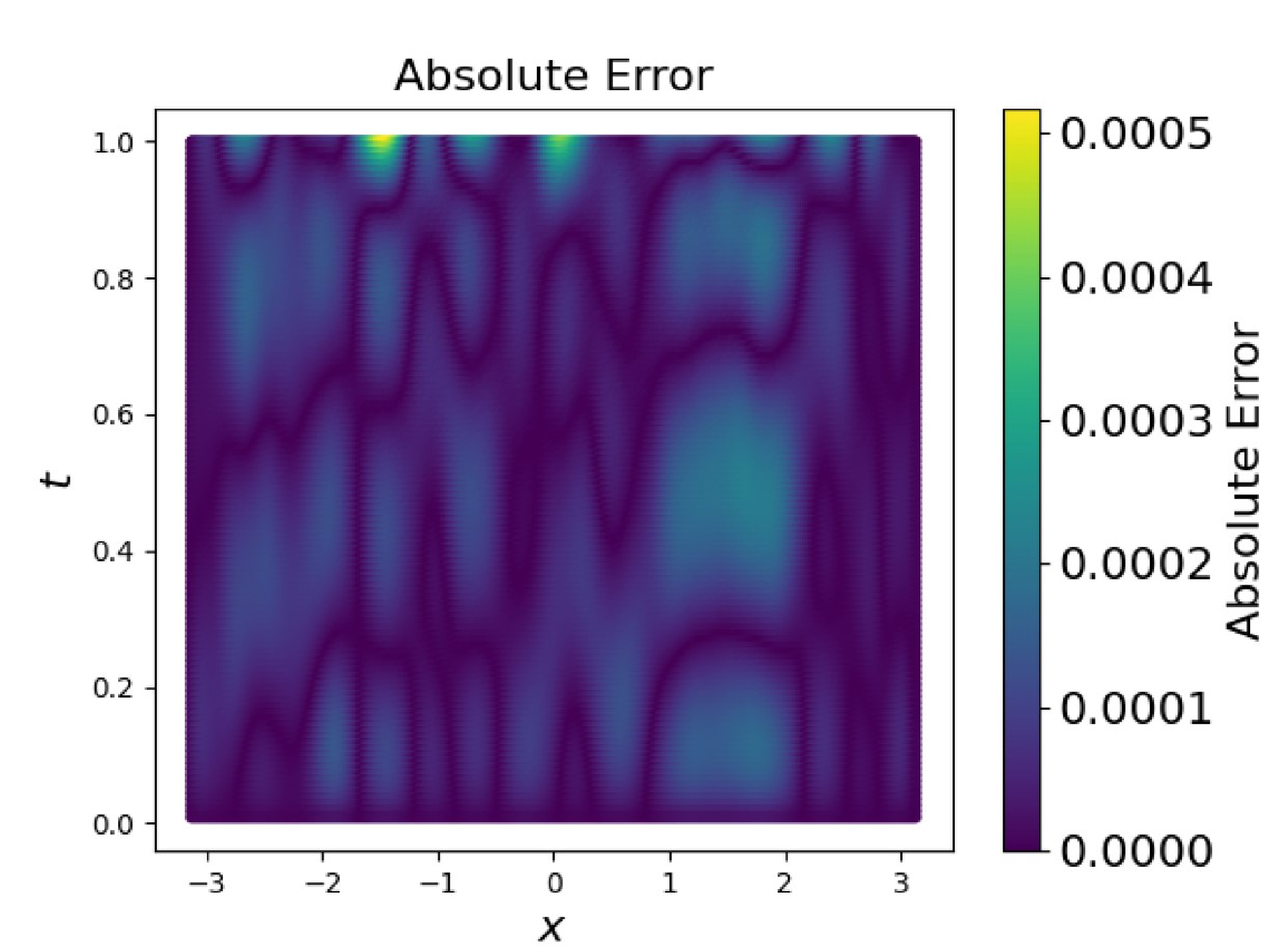}}
\subfigure[error, Sqr-ResNet]{ \label{Ex4_1DDiffRea_50x10_err_r}
\includegraphics[width=1.95in]{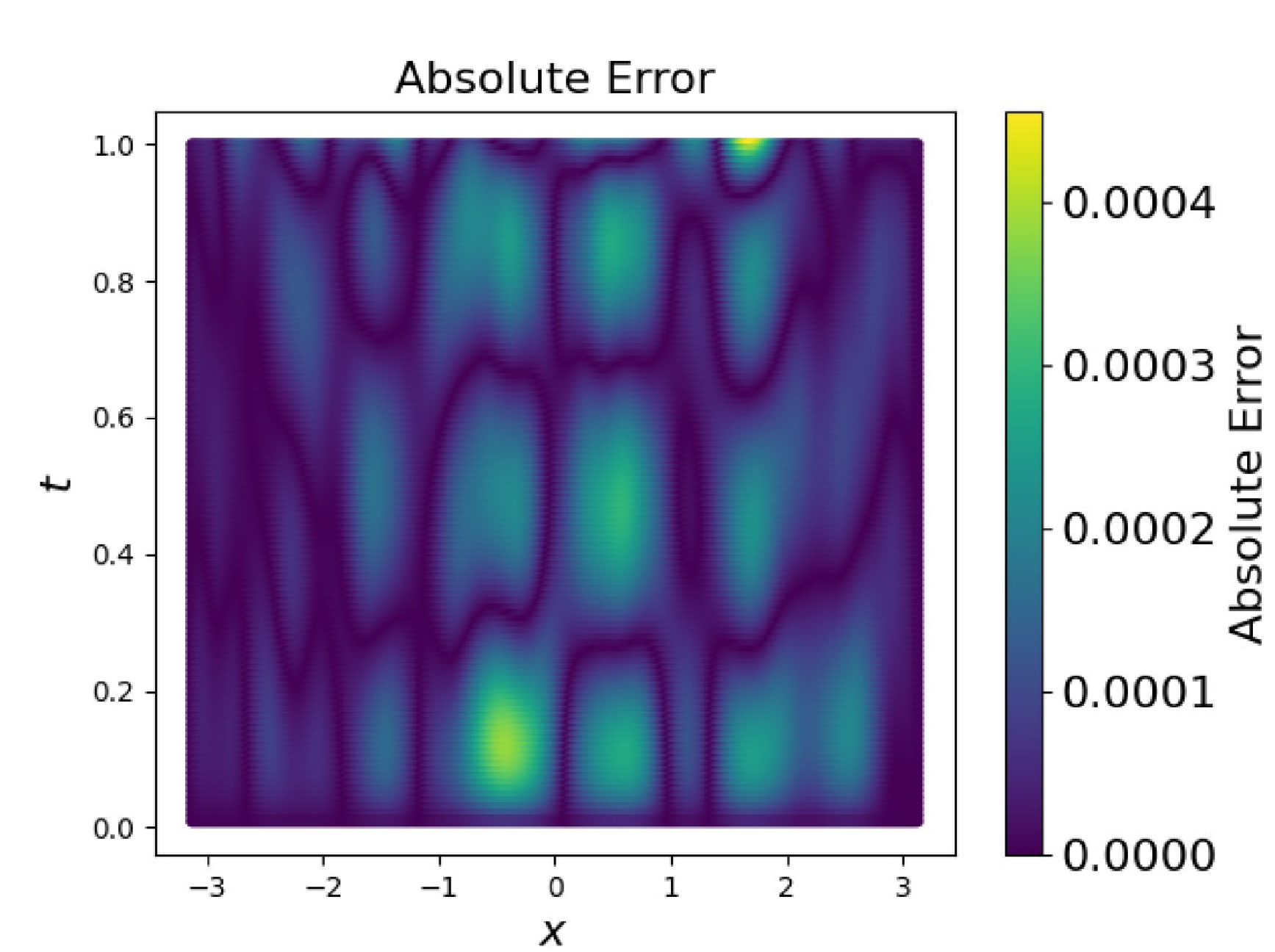}}
\caption{Example 2: Diffusion-reaction equation solution with NL=10.} \label{Ex4}
\end{figure}

Table~\ref{tab:ex4} compiles the results obtained using Plain network, Simple Highway network, and Sqr-ResNet with respect to NL (number of layers). It is evident that Plain Net has only successfully completed two cases, namely NL=5 and 15. Conversely, \rn exhibits improved performance but encounters failure at NL=30. Notably, \srn demonstrates the best overall performance by completing the training, achieving high accuracy, and maintaining stable performance.
Figure~\ref{Ex4} illustrates the loss and metric of NL=10 over the course of the training steps (top panel) and the maximum absolute error (bottom panel) corresponding to the smallest error during training. Analyzing Figure~\ref{Ex4_1DDiffRea_50x10_loss_p}, we observe divergence after 1000 steps and large absolute error in Figure~\ref{Ex4_1DDiffRea_50x10_err_p}. On the other hand both residual based methods, \rn and \srnc exhibit robust performance  on training data and small absolute error over validation data. 

}
\end{example}


\begin{example}\label{e6}\rm{
We aim to solve the heat equation given by:
\begin{equation}
    \frac{\partial u}{\partial t} = \alpha \frac{\partial^2 u}{\partial {x}_1^2}, \quad {x}_1 \in [0, 1], \quad t \in [0, 1]
\end{equation}
where \(\alpha = 0.4\) is the thermal diffusivity constant.
The boundary conditions are:
\begin{equation}
    u(0,t) = u(1,t) = 0.
\end{equation}
The initial condition is a sinusoidal function:
\begin{equation}
    u({x}_1, 0) = \sin\left(\frac{n \pi {x}_1}{L}\right), \quad 0 \leq {x}_1 \leq 1, \quad n = 1, 2, \ldots .
\end{equation}
The exact solution to the heat equation is:
\begin{equation}
    u({x}_1, t) = e^{-\frac{n^2 \pi^2 \alpha t}{L^2}} \sin\left(\frac{n \pi {x}_1}{L}\right).
\end{equation}
Here, \(L = 1\) represents the length of the domain, and \(n = 1\) corresponds to the frequency of the initial sinusoidal condition.

\begin{table}[!h]
  \captionof{table}{Example 3: The heat equation results.}\label{tab:ex6}
  \begin{center}
   \resizebox{\textwidth}{!}{
    \begin{threeparttable}
      \begin{tabular}{|c|ccl|ccl|ccl|} \hline
  & \multicolumn{3}{c|}{Plain Net}  
  & \multicolumn{3}{c|}{Simple HwNet}
  & \multicolumn{3}{c|}{Sqr-ResNet}
   \\
         NL&train loss &error &status       
   &train loss &error &status
   &train loss &error &status  \\ \hline   
5    
&1.0e-6 &7.0e-4 &completed   
&5.0e-7  &5.3e-4 &completed   
&4.8e-7  &4.2e-4 &completed \\\hline 

10   
&3.6e-6 &1.9e-3 &diverged\tnote{*}  
&5.8e-7  &9.4e-4 &completed   
&6.5e-7  &4.9e-4 &completed \\\hline 

15   
&5.2e-6 &3.5e-3 &diverged\tnote{\dag}     
&6.7e-7  &5.9e-4 &completed   
&7.4e-7  &9.3e-4 &completed \\\hline  

20   
&1.2e-6 &7.9e-4 &completed   
&\textbf{2.3e-7}  &\textbf{2.6e-4} &\textbf{completed}   
&5.3e-7  &6.7e-4 &completed \\\hline 

30   
&\ding{55} &\ding{55} &not trained   
&8.5e-7  &7.8e-4 &completed   
&4.6e-7  &5.8e-4 &completed \\ 
\hline
      \end{tabular}
      \begin{tablenotes}
        \item[*]  Diverged after step = 1000.
        \item[\dag]  Diverged after step = 1000.
\      \end{tablenotes}
    \end{threeparttable}}
  \end{center}
\end{table}

\begin{figure}
\centering%
\subfigure[Plain Net]{ \label{Ex6_1DHeat_50x15_err_p}
\includegraphics[width=1.95in]{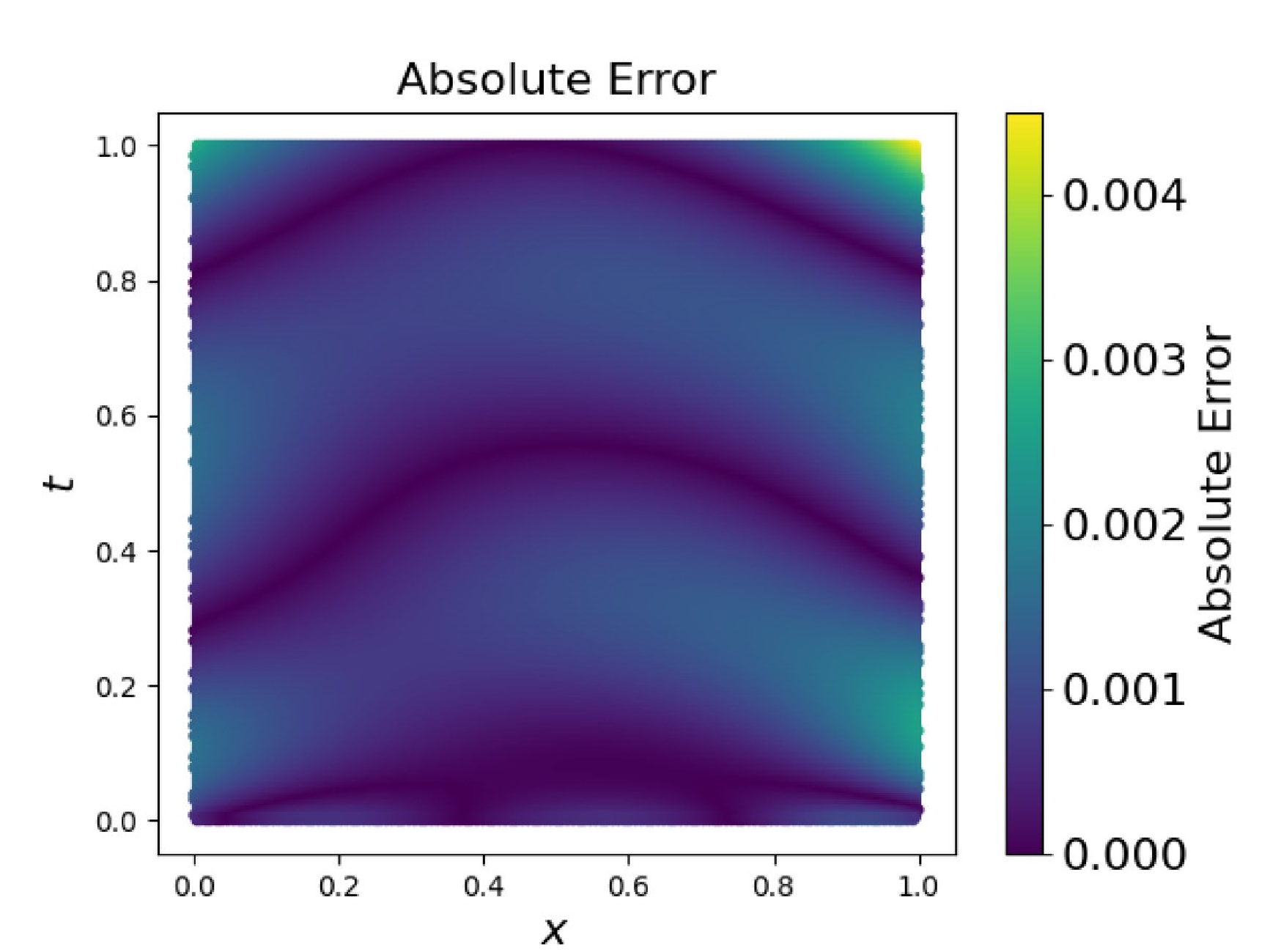}}
\subfigure[Simple HwNet]{ \label{Ex6_1DHeat_50x15_err_pr}
\includegraphics[width=1.95in]{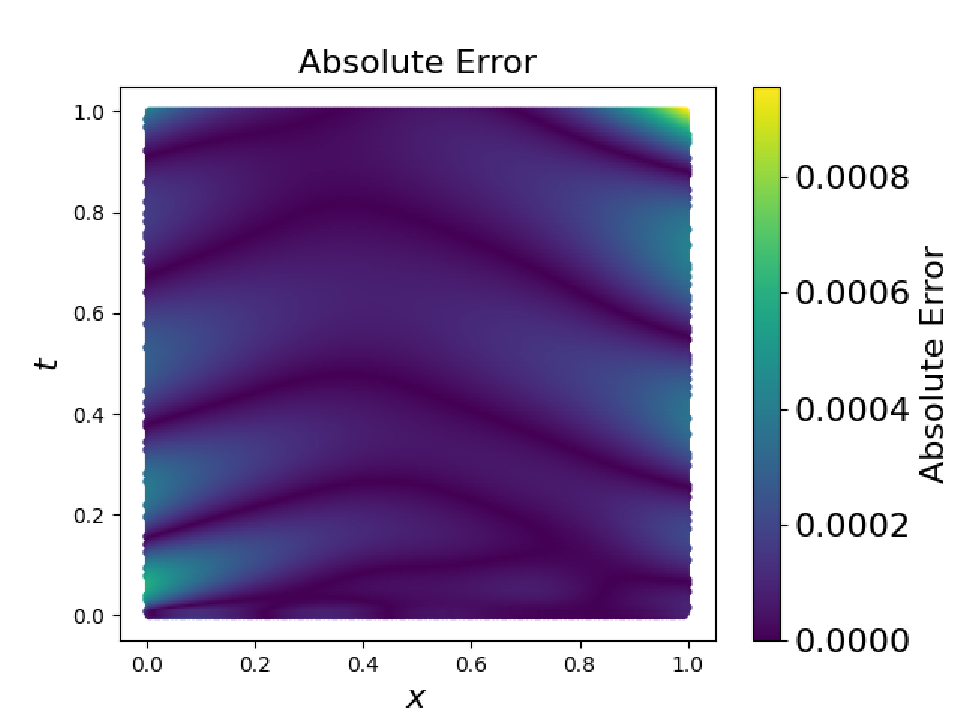}}
\subfigure[Sqr-ResNet]{ \label{Ex6_1DHeat_50x15_err_r}
\includegraphics[width=1.95in]{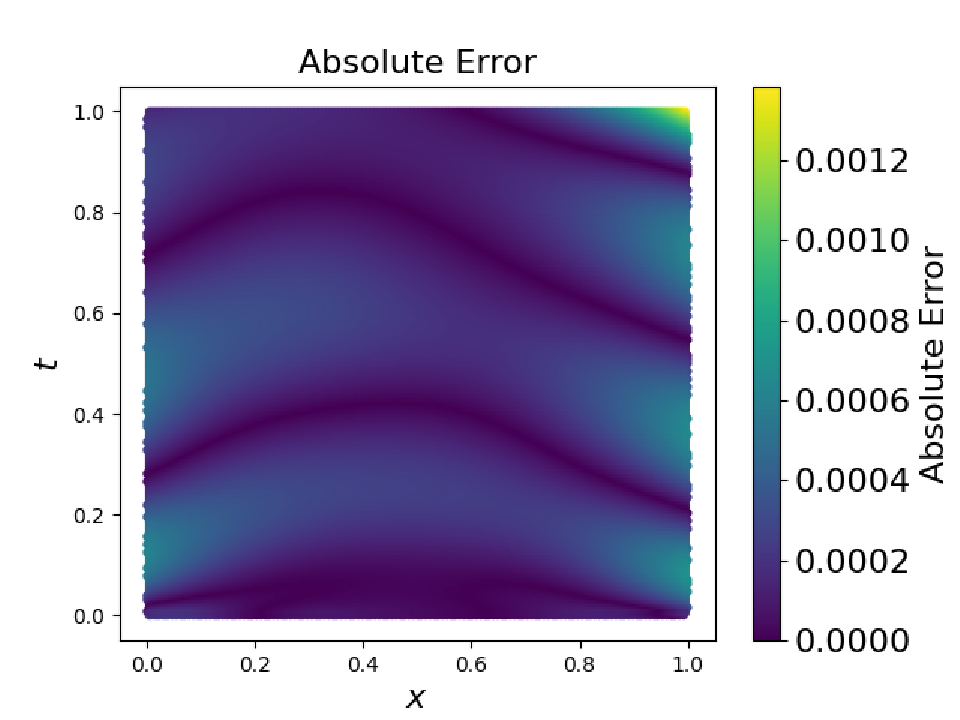}}
\caption{Example 3: The heat equation results with NL=15.} \label{Ex6}
\end{figure}

A total of 2780 collocation points, including 2540 interior, 80 boundary, and 160 initial points, are considered for this example. Table~\ref{tab:ex6} presents the results for Plain Net, simple HwNet, and Sqr-ResNet with respect to the NL. Plain Net completed 5000 steps only for NL=5 and 20, while NL=30 failed to train, producing large loss and error from the beginning of computation. Two cases, NL=10 and 15, produced large loss values after step=1000. On the other hand, both \rn and \srn completed the predefined number of iterations (5000) for all NL cases. However, the error for all cases with completed status is very close, with the best accuracy achieved for NL=20 using the simple highway network. 

Moreover, Figure~\ref{Ex6} presents the profile of the absolute error for the case with NL=15, using Plain Net, simple HwNet, and Sqr-ResNet. The figure showcases the best training results concerning the training loss value with respect to the steps. From these plots, it can be observed that \rn produces the smallest maximum absolute error, which occurred at step=5000. On the other hand, the plot shows the profile of the error for Plain Net at step=1000, which is very large compared with the other two methods due to incomplete training processes.

}
\end{example}

\begin{example}\label{e7}\rm{
In this example, we solve a nonlinear Schrödinger equation with periodic boundary conditions:
\begin{equation}
    i\frac{\partial u}{\partial t} + 0.5\frac{\partial^2 u}{\partial {x}_1^2} + |u|^2 u = 0, \quad {x}_1 \in [-5, 5], \quad t \in [0, \pi/2]. \label{eq:schrodinger}
\end{equation}
The initial condition is:
\begin{equation}
    u({x}_1,0) = 2 \text{sech}({x}_1),
\end{equation}
while the boundary conditions are:
\begin{equation}
    u(-5,t) = u( 5,t), \quad \frac{\partial u}{\partial {x}_1}(-5,t) = \frac{\partial u}{\partial {x}_1}(5,5).
\end{equation}
The exact solution is referenced from \cite{Raissi19}.
Let \( h \) denote the real part and \( v \) denote the imaginary part of the solution \( u \). We place a multi-output neural network prior on \( u(x_1,t) = [h(x_1,t), v(x_1,t)] \).

\begin{table}[!h]
  \captionof{table}{Example 4: Nonlinear Schrödinger equation   error $(\epsilon)$ for real $(h)$ and imaginary $(v)$  parts.}\label{tab:ex7}
  \begin{center}
   \resizebox{\textwidth}{!}{
    \begin{threeparttable}
      \begin{tabular}{|c|cccl|cccl|cccl|} \hline
  & \multicolumn{4}{c|}{Plain Net}  
  & \multicolumn{4}{c|}{Simple HwNet}
  & \multicolumn{4}{c|}{Sqr-ResNet}
   \\
 NL&train loss & $\epsilon({ h})$& $\epsilon({ v})$ &status       
   &train loss & $\epsilon({ h})$& $\epsilon({ v})$ &status
   &train loss & $\epsilon({ h})$& $\epsilon({ v})$ &status  \\ \hline   
5    
&1.4e-4 &5.5e-2&9.2e-2 &completed   
&8.1e-5  &3.9e-2&6.6e-2 &completed   
&1.1e-5  &4.7e-3&7.1e-3 &completed \\\hline 

10   
&3.8e-5&6.8e-3 &1.1e-2 &completed   
&4.2e-5  &1.1e-2&1.9e-2 &completed 
&1.1e-5  &2.6e-3&3.4e-3 &completed \\ \hline 

15   
&\ding{55} &\ding{55} &\ding{55} &not trained   
&3.3e-5  &1.5e-2&2.6e-2 &completed   
&\textbf{6.7e-6}  &\textbf{1.9e-3}&\textbf{2.6e-3} &\textbf{completed} \\\hline  

20   
&1.1e-4 &2.6e-2&4.4e-2 &completed   
&4.2e-5  &1.3e-2&2.2e-2 &completed   
&3.5e-5  &5.7e-3&9.6e-3 &completed \\\hline 

30   
&2.0e-3 &3.1e-1&5.0e-1 &completed   
&6.9e-5  &1.5e-2&2.4e-2 &completed   
&3.1e-5  &1.2e-2&2.2e-3 &completed \\ 
\hline
      \end{tabular}
    \end{threeparttable}}
  \end{center}
\end{table}

\begin{figure}
\centering%
\subfigure[Plain Net]{ \label{Ex7_1Dschro_50x30_loss_p}
\includegraphics[width=1.95in]{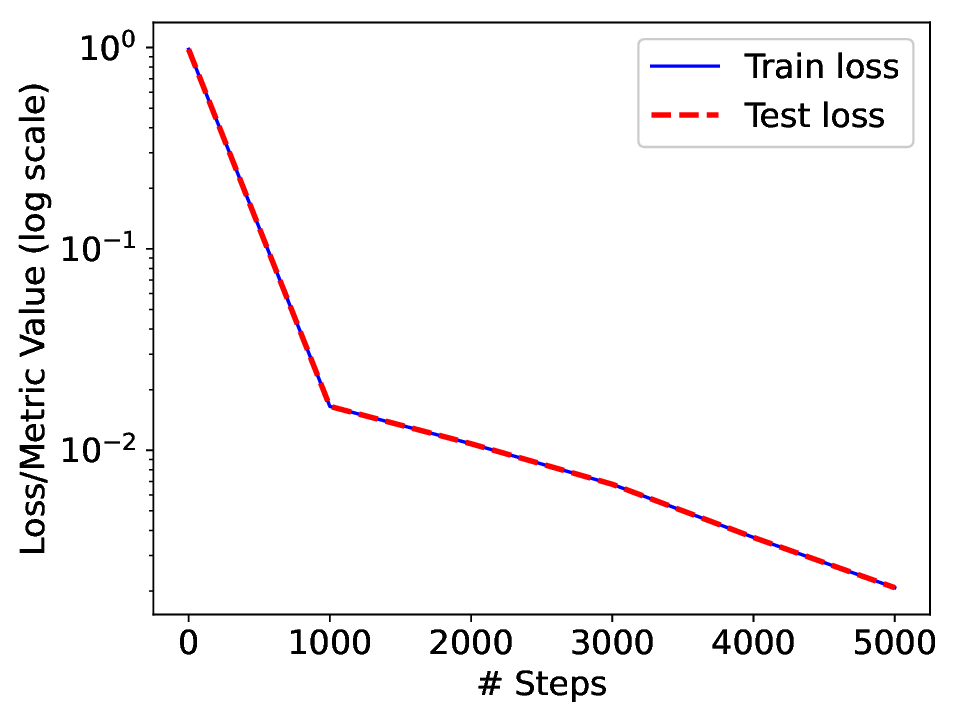}}
\subfigure[Simple HwNet]{ \label{Ex7_1Dschro_50x30_loss_pr}
\includegraphics[width=1.95in]{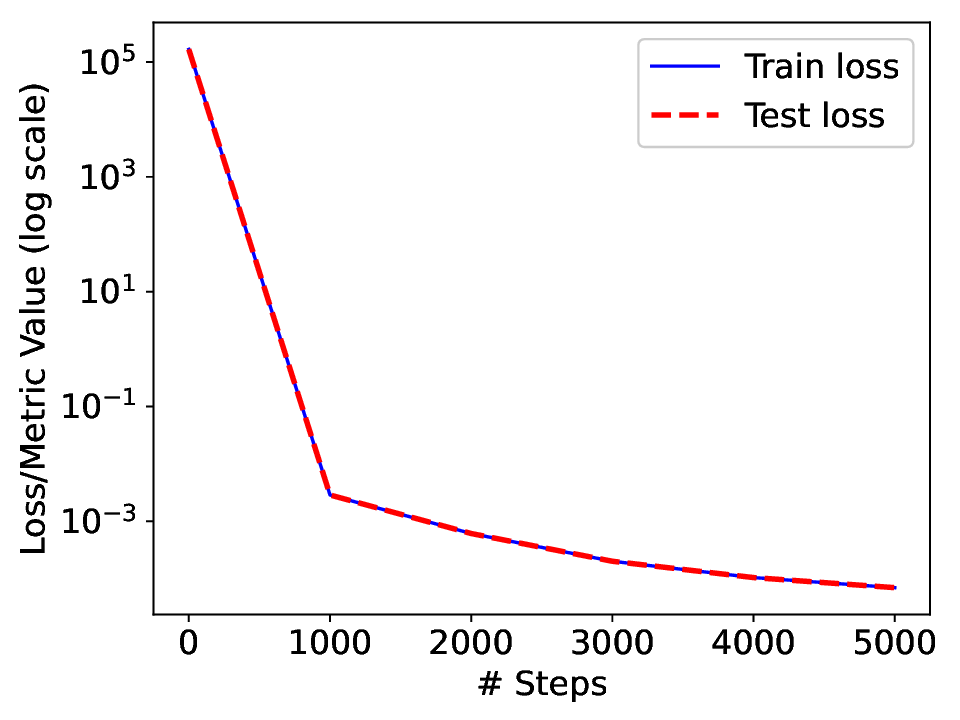}}
\subfigure[Sqr-ResNet]{ \label{Ex7_1Dschro_50x30_loss_r}
\includegraphics[width=1.95in]{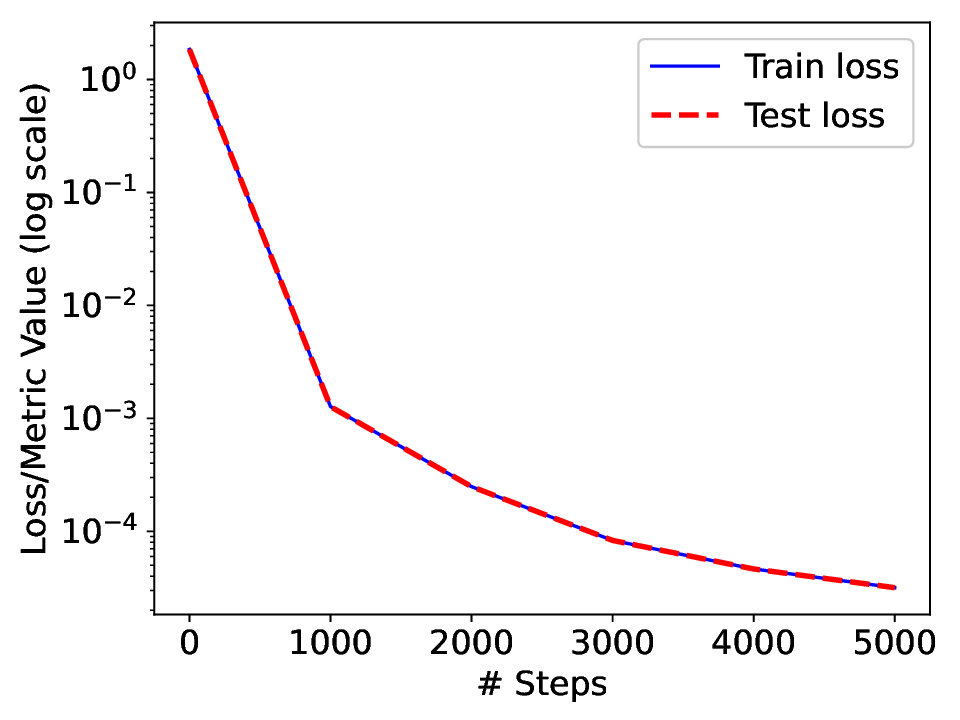}}
\caption{Example 4: Nonlinear Schrödinger equation results with NL=30.} \label{Ex7}
\end{figure}

\noindent
In this example, 10000 interior points, 20 boundary points, and 200 points for the initial condition are considered. Table~\ref{tab:ex7} presents the train loss. It also lists the error on the real and imaginary terms shown as $\epsilon({\rm h})$ and $\epsilon({\rm v})$. In this example, Plain Net behaves more stably than in previous examples in terms of completing the training process. However, still at NL=15, a divergence after step=1000 occurred. From this table we can see that there is a rise in errors at NL=30. The residual-based method, simple highway network, behaves more stably by keeping the order of error magnitudes at -2. However, the best performance can be seen for \srnc where the smallest errors compared to the two other architectures are observed, and the results are stable around the magnitude of order -3 as NL increases. Note that the best accuracy happened at NL=15 using \srnd Figure~\ref{Ex7} shows the loss and metric values with respect to the step. Clearly, \srn leads to the best training performance in terms of accuracy and convergence, and Plain Net is the worst one.

}
\end{example}

\begin{example}\label{e8}\rm{
In the final example, we solve a 2D linear elasticity problem in solid mechanics. The governing equations are:
\begin{align}
    \frac{\partial \sigma_{{x}_1{x}_1}}{\partial {x}_1} + \frac{\partial \sigma_{{x}_1{x}_2}}{\partial {x}_2} + f_{x_1} &= 0, \quad {x}_1, {x}_2 \in [0, 1], \\
    \frac{\partial \sigma_{{x}_1{x}_2}}{\partial {x}_1} + \frac{\partial \sigma_{{x}_2{x}_2}}{\partial {x}_2} + f_{x_2} &= 0.
\end{align}
The linear elastic constitutive relations are:
\begin{align}
    \sigma_{{x}_1{x}_1} &= (\lambda + 2\mu)\epsilon_{{x}_1{x}_1} + \lambda\epsilon_{{x}_2{x}_2}, \\
    \sigma_{{x}_2{x}_2} &= (\lambda + 2\mu)\epsilon_{{x}_2{x}_2} + \lambda\epsilon_{{x}_1{x}_1}, \\
    \sigma_{{x}_1{x}_2} &= 2\mu\epsilon_{{x}_1{x}_2},
\end{align}
where the strains are defined as:
\begin{align}
    \epsilon_{{x}_1{x}_1} &= \frac{\partial u_{{x}_1}}{\partial {x}_1}, \\
    \epsilon_{{x}_2{x}_2} &= \frac{\partial u_{{x}_2}}{\partial {x}_2}, \\
    \epsilon_{{x}_1{x}_2} &= \frac{1}{2}\left(\frac{\partial u_{{x}_1}}{\partial {x}_2} + \frac{\partial u_{{x}_2}}{\partial {x}_1}\right).
\end{align}
The body forces applied are:
\begin{align}
    f_{x_1} &= \lambda\left[4\pi^2\cos(2\pi {x}_1)\sin(\pi {x}_2) - \pi\cos(\pi {x}_1)Q{x}_2^3\right] \nonumber \\
    &\quad + \mu\left[9\pi^2\cos(2\pi {x}_1)\sin(\pi {x}_2) - \pi\cos(\pi {x}_1)Q{x}_2^3\right], \\ 
    f_{x_2} &= \lambda\left[-3\sin(\pi {x}_1)Q{x}_2^2 + 2\pi^2\sin(2\pi {x}_1)\cos(\pi {x}_2)\right] \nonumber \\
    &\quad + \mu\left[-6 \sin(\pi {x}_1)Q{x}_2^2 + 2 \pi^2\sin(2\pi {x}_1)\cos(\pi {x}_2) + \pi^2\sin(\pi {x}_1)Q{x}_2^4/4\right],
\end{align}
The displacement boundary conditions are:
\begin{align}
    u_{x_1}({x}_1, 0) &= u_{x_1}({x}_1, 1) = 0, \\
    u_{x_2}(0, {x}_2) &= u_{x_2}(1, {x}_2) = u_{x_2}({x}_1, 0) = 0.
\end{align}
The traction boundary conditions are:
\begin{align}
    \sigma_{{x}_1{x}_1}(0, {x}_2) &= \sigma_{{x}_1{x}_1}(1, {x}_2) = 0,\\
    \sigma_{{x}_2{x}_2}({x}_1, 1) &= (\lambda + 2\mu)Q\sin(\pi {x}_1).
\end{align}
We use the following parameters: \(\lambda = 1\), \(\mu = 0.5\), and \(Q = 4\). The exact solutions are \(u_{x_1}({x}_1, {x}_2) = \cos(2\pi {x}_1)\sin(\pi {x}_2)\) and \(u_{x_2}({x}_1, {x}_2) = Q{x}_2^4/4 \sin(\pi {x}_1)\).

\begin{table}[!h]
  \captionof{table}{Example 5: Elastostatic equation with NN=200.}\label{tab:ex8}
  \begin{center}
   \resizebox{\textwidth}{!}{
    \begin{threeparttable}
      \begin{tabular}{|c|ccl|ccl|ccl|} \hline
  & \multicolumn{3}{c|}{Plain Net}  
  & \multicolumn{3}{c|}{Simple HwNet}
  & \multicolumn{3}{c|}{Sqr-ResNet}
   \\
         NL&train loss &error &status       
   &train loss &error &status
   &train loss &error &status  \\ \hline   
5    
&3.6e-5 &3.2e-3 &completed   
&1.1e-4  &9.2e-3 &completed   
&1.9e-5  &2.6e-3 &completed \\\hline 

10   
&5.7e-5 &4.1e-3 &completed   
&2.5e-5  &2.8e-3 &completed   
&\textbf{4.1e-5}  &\textbf{2.7e-3} &\textbf{completed} \\\hline 

15   
&1.9e-4 &6.5e-3  &completed
&5.7e-5  &4.5e-3 &completed   
&8.1e-5  &4.8e-3 &completed \\\hline  

20   
&\ding{55} &\ding{55} &not trained   
&6.7e-5  &3.5e-3 &completed   
&1.5e-4  &7.1e-3 &completed \\\hline 

30   
&\ding{55} &\ding{55} &not trained   
&3.7e-4  &1.2e-2 &completed   
&7.9e-5  &5.8e-3 &completed \\ 
\hline
      \end{tabular}
    \end{threeparttable}}
  \end{center}
\end{table}

In this example, we consider 500 interior points and 500 points on the boundary. The results presented in Table~\ref{tab:ex8} indicate a decrease in the performance of Plain Net as NL increases, with the last two NL values (NL=20 and 30) not being trained successfully. On the contrary, the residual-based methods exhibit better performance, with all cases successfully implemented. Notably, \srn maintains the error's order of magnitude at -3, showcasing a more stable computation across various NL. Additionally, the smallest error is observed at NL=10 using \srnd

}
\end{example}


\section{Conclusions}

In this study, we introduced two new types of network architectures: the Simple Highway Network and the Squared Residual Network. These are designed to make solving partial differential equations  with deep learning more stable.
Our contributions emphasize the enhanced final results and the underlying reasons for the superior performance of our proposed networks, which are applicable for evaluating the performance of other architectures.

Our findings show that the Squared Residual Network significantly improves stability during weight updates for function approximation example. Smoother and appropriately scaled weight updates lead to better training accuracy and quicker convergence. Conversely, large weight updates often result in higher errors. Our residual-based architectures also effectively reduce the gradient of loss related to weight updates, improving training effectiveness.

Physics-informed neural networks rely on accurate function approximation. Our networks excel in solving a wide range of PDEs, whether they are time-dependent, independent, linear, or nonlinear. The Squared Residual Network, especially, consistently performs well. It successfully completes training where plain networks fail, achieving high accuracy.
Conversely, while a plain network may demonstrate that deeper architectures can lead to instability and reduced accuracy, it has also been shown that deeper networks can achieve better accuracy when an appropriate architecture is selected. 
The Squared Residual Network's stability and accuracy improvements make it a promising tool for advancing deep learning in PDEs and computational physics.


%

\end{document}